%% file: main.tex
\documentclass[10pt,twocolumn,letterpaper]{article}

\usepackage[pagenumbers]{iccv} %

\input{preamble}
\definecolor{kleinblue}{RGB}{0, 47, 167} 
\definecolor{kleinred}{HTML}{bc1919}

\usepackage[pagebackref,break links,colorlinks,linkcolor=kleinred, citecolor=kleinblue]{hyperref}
\usepackage{url}
\usepackage{booktabs}
\usepackage{siunitx}
\usepackage{pifont} %
\usepackage{graphicx}
\usepackage{xspace}
\usepackage{adjustbox}
\usepackage{tablefootnote}
\usepackage{threeparttable}
\usepackage{booktabs}
\usepackage{longtable}
\usepackage{multirow}

\newcommand{\ourmethodname}[1]{OCCAM#1}

\newcommand{\tick}{\ding{51}}
\newcommand{\cross}{\ding{55}}
\newcommand{\numobj}[0]{K}
\newcommand{\maxobj}[0]{$K_{\operatorname{max}}$}
\newcommand{\numcls}[0]{C}
\newcommand{\cropresize}[0]{``Gray BG + Crop''}
\newcommand{\citeobjectdiscovery}[0]{\citep{SlotAttention, SlotDiffusion, Dinosaur, ft-dinosaur, SAVI, SAVI++, greff2019multi}}

\definecolor{byzantium}{rgb}{0.44, 0.16, 0.39}
\definecolor{cardinal}{rgb}{0.77, 0.12, 0.23}
\definecolor{cadetblue}{rgb}{0.37, 0.62, 0.63}

\newcommand{\highlightgreen}[1]{{\color{PineGreen}{#1}\color{black}}}
\newcommand{\highlightred}[1]{{\color{cardinal}{#1}\color{black}}}
\newcommand{\highlightpurple}[1]{{\color{byzantium}{#1}\color{black}}}
\newcommand{\highlightblue}[1]{{\color{cadetblue}{#1}\color{black}}}

\usepackage[fixed]{fontawesome5}
\def\paperID{4140} %
\def\confName{ICCV}
\def\confYear{2025}

\title{\vspace{-0.75cm}Are We Done with Object-Centric Learning?}

\author{Alexander Rubinstein \qquad Ameya Prabhu \qquad Matthias Bethge \qquad Seong Joon Oh\vspace{0.1cm}\\
T\"ubingen AI Center, University of T\"ubingen\vspace{0.2cm}\\
{\normalsize \raisebox{-1pt}{\faGlobe} \href{https://alexanderrubinstein.github.io/are-we-done-with-ocl/ }{Project Page} \quad  \raisebox{-1pt}{\faGithub} \href{https://github.com/AlexanderRubinstein/OCCAM}{OCCAM Codebase}}
}

\begin{document}
\maketitle
\input{sec/0_abstract}    
\input{sec/1_intro}
\input{sec/2_background}

\input{sec/3_method}

\input{sec/4_experiments}
\input{sec/5_discussions}
{
    \small
    \bibliographystyle{ieeenat_fullname}
    \bibliography{main}
}

\appendix
\input{sec/X_suppl}

\end{document}

%% file: sec/0_abstract.tex
\begin{abstract}
Object-centric learning (OCL) seeks to \textit{learn} representations which only encode an object, isolated from other objects or background cues in a scene. This approach underpins various aims, including out-of-distribution (OOD) generalization, sample-efficient composition, and modeling of structured environments. Most research has focused on developing unsupervised mechanisms that separate objects into discrete slots in the representation space, evaluated using unsupervised object discovery. However, with recent sample-efficient segmentation models, we can separate objects in the pixel space and encode them independently. This achieves remarkable zero-shot performance on OOD object discovery benchmarks, is scalable to foundation models, and can handle a variable number of slots out-of-the-box. Hence, the goal of OCL methods to obtain object-centric representations has been largely achieved. Despite this progress, a key question remains: How does the ability to separate objects within a scene contribute to broader OCL objectives, such as OOD generalization? We address this by investigating the OOD generalization challenge caused by spurious background cues through the lens of OCL. We propose a novel, training-free probe called \textbf{Object-Centric Classification with Applied Masks} (\textbf{\ourmethodname{}}), demonstrating that segmentation-based encoding of individual objects significantly outperforms slot-based OCL methods. However, challenges in real-world applications remain. We provide the toolbox for the OCL community to use scalable object-centric representations, and focus on practical applications and fundamental questions, such as understanding object perception in human cognition.  Our code is available \href{https://github.com/AlexanderRubinstein/diverse-universe-public}{here}.
\end{abstract}

%% file: sec/1_intro.tex
\section{Introduction}
\label{sec:intro}

\begin{figure}[t]
    \vspace{-0.4cm}
    \includegraphics[width=1.05\columnwidth]{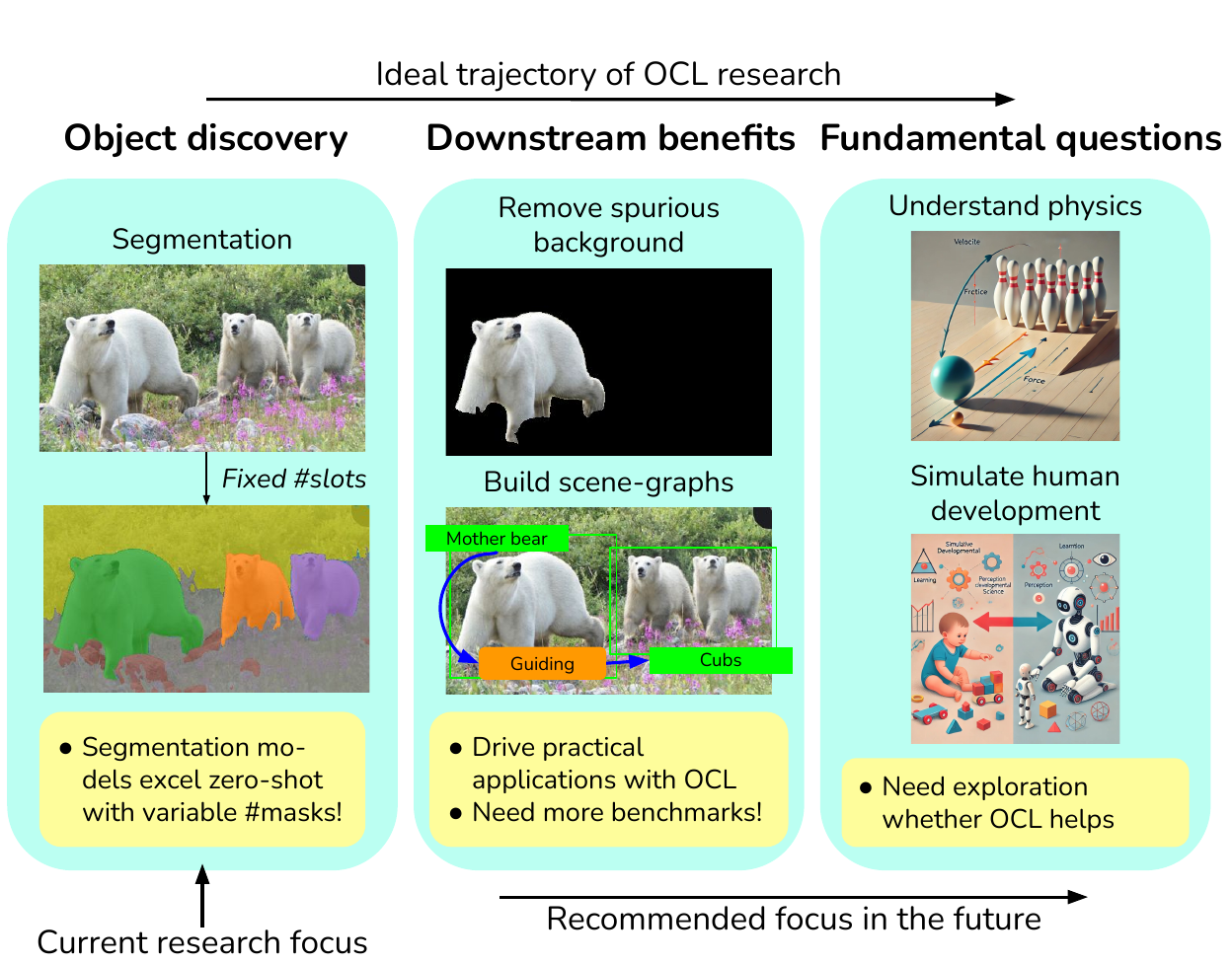}
    \caption{\textbf{Where Should We Go?} Object-centric learning (OCL) has focused on developing unsupervised mechanisms to separate the representation space into discrete \textit{slots}. However, the inherent challenges of this task have led to comparatively less emphasis on exploring downstream applications and exploring fundamental benefits. Here, we introduce simple, effective OCL mechanisms by separating objects in pixel space and encoding them independently. We present a case study that demonstrates the downstream advantages of our approach for mitigating spurious correlations. We outline the need to develop benchmarks aligned with fundamental goals of OCL, and explore the downstream efficacy of OCL representations.}
    \label{fig:overview}
\end{figure}

Object-centric learning (OCL) seeks to develop representations of complex scenes that independently encode each foreground object separately from background cues, ensuring that one object’s representation is not influenced by others or the background \citep{greff2019multi, MONET}. This constitutes a foundational element for many objectives: it supports modeling of structured environments \citep{ScholkopfCausal}, enables robust out-of-distribution (OOD) generalization \citep{OCL-OOD-GEN-SEGMENT, CoBalT, wiedemer2024provable, Ferdinand1, Ferdinand2}, facilitates compositional perception of complex scenes \citep{CompositionalNatureOfScenes}, and deepens our understanding of object perception in human cognition \citep{CognitionSpelke, InfantsCognition, CognitionOxford}. However, despite these broad goals, most research in OCL has centered on advancing ``slot-centric” methods that separate objects and encode them into slots, evaluated using unsupervised object discovery as the primary metric \citeobjectdiscovery. In this paper, we challenge the continued emphasis on developing mechanisms to separate objects in representation space as the main challenge to be addressed in OCL.\vspace{0.1cm}

\noindent We first show that sample-efficient class-agnostic segmentation models, such as High-Quality Entity Segmentation (HQES) \citep{EntitySeg} are far better alternatives to the latest slot-centric OCL approaches, already achieving impressive zero-shot object discovery. Moreover, these models are scalable, with foundation models like Segment Anything (SAM) \citep{SAM, SAMv2} showing remarkable zero-shot segmentation, addressing much of what is usually tackled with slot-centric approaches. Yet, the broader potential of OCL remains largely unexplored. We pose a critical question: How does the ability to separate objects within scenes contribute to other OCL objectives, such as OOD generalization?\vspace{0.1cm}

\noindent We bridge this gap by directly linking OCL to OOD generalization, especially in known hard settings with spurious background cues. We introduce \textbf{Object-Centric Classification with Applied Masks} (\textbf{\ourmethodname{}}), a simple, object-centric probe for robust zero-shot image classification. \ourmethodname{} consists of two stages: (1) generating object-centric representations via object-wise mask generation, and (2) applying OCL representations to downstream applications, such as image classification in the presence of spurious backgrounds, by selectively focusing on relevant object features while discarding misleading background cues.\vspace{0.1cm}

\noindent Empirically, we find that, on Stage (1), sample-efficient segmentation models outperform current OCL approaches in obtaining object-centric representations without additional training. 
However, on Stage (2) — the task of identifying relevant object cues amidst numerous possible masks — remains a challenge. Nevertheless, when Stage (2) is executed correctly, simple OCL probes such as \ourmethodname{} already have the potential for robust OOD generalization.\vspace{0.1cm}

\noindent We recommend more focus by future OCL works on creating benchmarks, methodologies for testing real-world applications where object-centric representations offer clear practical benefits, encouraging theory motivated by specific real-world tasks, and exploring fundamental questions, such as how object perception works in human cognition. 

%% file: sec/2_background.tex
\section{Related work}
\label{sec:related_work}

We cover prior work in the object-centric learning (OCL) community from three different angles: motivation, evaluation, and methodologies.\vspace{0.1cm} 

\noindent \textbf{Motivation for OCL.} The OCL community has inspired research from different perspectives. From one perspective, learning object-centric representations can help discover latent variables of the data-generating process, such as object position and color \citep{LatentFactorsRecoverRealWorld}, or even identify its causal mechanisms \citep{liu2023causal, ScholkopfCausal} by encoding structural knowledge that allows interventions and changes. From another perspective, OCL aims to simulate human cognition \citep{CognitionSpelke, InfantsCognition, CognitionOxford} in neural networks. For example, infants intuitively understand physics by tracking objects with consistent behavior over time \citep{OCL-OOD-GEN-SEGMENT}. They later reuse this knowledge to learn new tasks quickly. Advances in OCL can help neural networks develop this ability as well.
In addition to that, some studies focus on understanding the compositional nature of scenes \citep{CompositionalNatureOfScenes} by providing separate representations for different elements (e.g., human, hat, bed, table) and their interactions (a cat wearing a hat or a bear guiding cubs). Several papers claim that there is a potential to improve sample efficiency \citep{Ferdinand2} and generalization \citep{SlotAttention, SAVI, Dinosaur, wiedemer2024provable, Ferdinand1, Ferdinand2} or object-centric methods can be more robust \citep{Dinosaur, CoBalT}. Others refer to the structure of the world, saying that the fundamental structure of the physical world is compositional and modular \citep{SlotDiffusion} or that humans understand the world in terms of separate objects \citep{SAVI, ft-dinosaur}. However, we have observed a consistent lack of empirical evidence demonstrating that object-centric approaches improve sample efficiency or aid in identifying causal mechanisms. To address this gap, we believe more empirical research is needed. As a first step, we show that robust classification is achievable even in the presence of explicitly distracting backgrounds and other object interference.

\noindent \textbf{OCL evaluation.} Measuring progress on the primary motivations of object-centric learning is a hard problem and suffers from a chronic lack of scalable benchmarks. Hence, empirical support for the commonly claimed benefits, such as parameter/learning efficiency \citep{SAVI, Ferdinand2} and improved generalization \citep{OCL-OOD-GEN-SEGMENT, CoBalT, wiedemer2024provable, Ferdinand1, Ferdinand2} or better understanding of representations, remains limited. 
Some papers study the link between object-centric learning and downstream applications. These include reinforcement learning \citep{Watters2019COBRADM, Kulkarni2019UnsupervisedLO, Berner2019Dota2W, sun2018predicting, yoon2023investigation}, scene representation and generation \citep{Kulkarni2019UnsupervisedLO, ElNouby2018TellDA, Matsumori2021UnifiedQT, MONET}, reasoning \citep{webb2023systematic, Yang2020ObjectCentricDO}, and planning \citep{migimatsu2019objectcentric}. We highlight that these papers provide a valuable contribution to benchmarking progress in the OCL field. However, most research does not focus on these tasks. Much of the progress is tracked by unsupervised object discovery benchmarks, essentially entity segmentation \citeobjectdiscovery. Model performance is usually quantified with foreground adjusted random index (FG-ARI) \citep{FG_ARI_1971, FG_ARI_1985, SAVI}, which is a permutation-invariant clustering metric or mean best overlap (mBO) \citep{mbo_metric, Dinosaur}. These evaluations primarily assess whether slots reliably isolate individual objects — a criterion we argue is overly restrictive in the broader context of object-centric learning. In our paper, we urge more work to additionally evaluate downstream applications, particularly given the emergence of foundational segmentation models that significantly outperform object-centric methods on standard object discovery tasks (see Table~\ref{tab:quant_object_discovery} and Figure~\ref{tab:segmentation_qualitative_movi_e}).\vspace{0.1cm}

\begin{figure*}[ht]
\vspace{-0.3cm}
    \centering
    \includegraphics[width=1.0\textwidth]{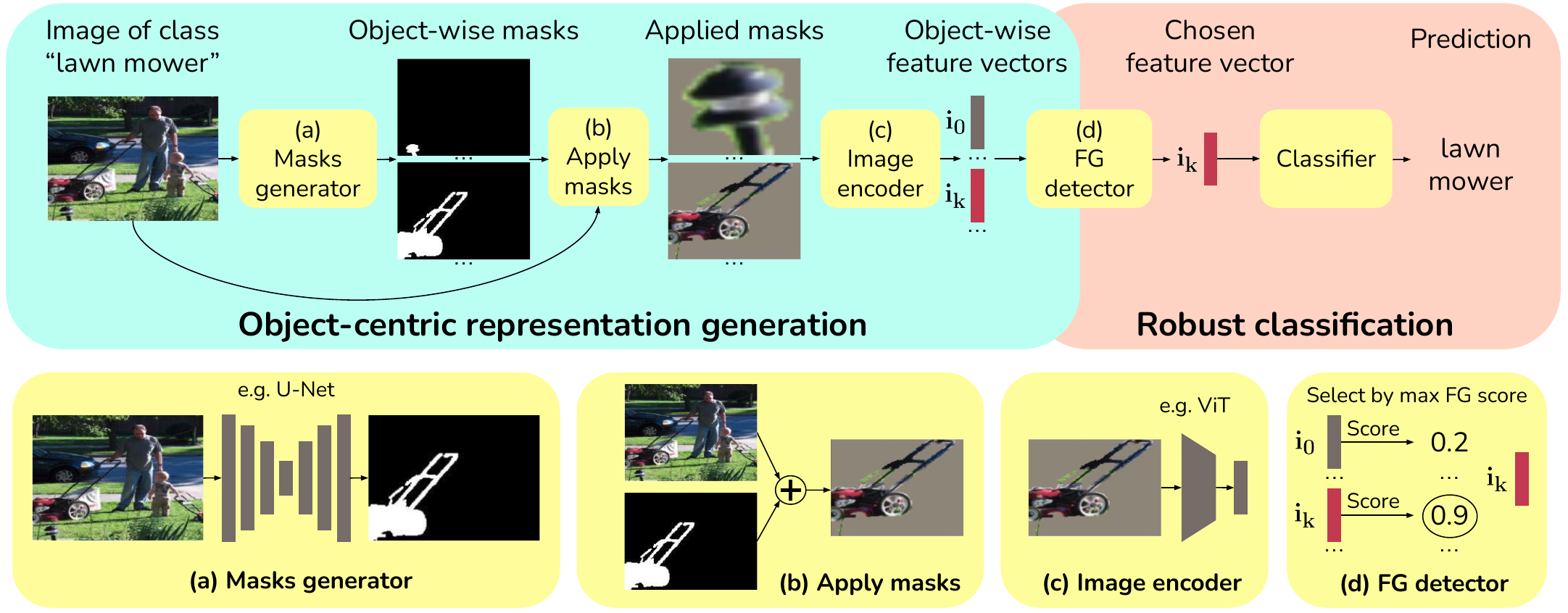}
    \caption{\textbf{Overview of Object-Centric Classification with Applied Masks (\ourmethodname{})}. There are two main parts. The first part (§~\ref{sec:method:generate_representations}) uses entity segmentation masks for \textbf{object-centric representation generation}. The second part (§~\ref{sec:method:robust_classifier}) performs \textbf{robust classification} by selecting representations corresponding to the foreground object and using them for classification. Indices $[i_0, \ldots, i_k, \ldots]$ correspond to each object in the scene.}
    \label{fig:method_figure}
\end{figure*}

\noindent \textbf{OCL methodologies.} OCL captured widespread attention with the introduction of SlotAttention \citep{SlotAttention}, which enabled iterative learning of separate latent representations for each object in an image. These latent “slots” can then be decoded back to the pixel space. Extensions have included SlotAttention paired with diffusion decoders \citep{SlotDiffusion} and SlotAttention architectures built on top of DINO \citep{Dinosaur, ft-dinosaur} features. Dinosaur \citep{Dinosaur} uses pre-trained self-supervised DINO \citep{DINO} features as a target for reconstruction loss. This loss is used to train a decoder with Slot Attention \citep{SlotAttention} on top of the ResNet \citep{ResNet} encoder. FT-Dinosaur \citep{ft-dinosaur} improves Dinosaur by replacing the ResNet encoder with a DINO-ViT \citep{ViT} encoder separate from the one used to compute target features. It jointly fine-tunes the encoder with the decoder. SlotDiffusion \citep{SlotDiffusion} uses pre-trained features from the Stable Diffusion Encoder \citep{StableDiffusion} and trains a diffusion-based decoder with Slot Attention \citep{SlotAttention} on top of them. In video contexts, sequential adaptations leverage temporal dependencies \citep{SAVI} and depth information \citep{SAVI++}. Some studies also propose theoretical foundations for OCL \citep{wiedemer2024provable, brady2023provably}. There is also a line of work that studies object-centric representation in the context of out-of-distribution (OOD) generalization in segmentation \citep{OCL-OOD-GEN-SEGMENT}, compositional generalization \citep{wiedemer2024provable, Ferdinand1, Ferdinand2}, and classification, e.g., CoBalT \citep{CoBalT} that employs model distillation and slots clustering into concepts to refine feature quality. In our experiments, we compare with the latest methods --  SlotDiffusion \citep{SlotDiffusion} and (FT-)Dinosaur \citep{Dinosaur, ft-dinosaur} for object discovery and CoBalT \citep{CoBalT} across robust classification benchmarks.

%% file: sec/3_method.tex
\section{Method}
\label{sec:method}

This section gives an overview of our proposed method. Subsection §\ref{sec:method:notations} defines the notation needed for the method description in §\ref{sec:method:method}.

\subsection{Notations}
\label{sec:method:notations}

We denote an image as 
$ x\in[0, 1]^{[3, \text{ H}, \text{ W}]}$
and a label as $y \in \mathcal{Y} = \{1, \ldots, \numcls\}$, where $\numcls$ is the number of classes. We will write an image encoder, or a feature extractor, as $\psi$ and image embedding, or feature vector, as $\psi(x) \in \mathbb{R}^{d}$, where $d\geq 1$ is the feature dimensionality. 
We define the classifier's pre-softmax logits as $f(\psi(x)) \in \mathbb{R}^{|\mathcal{Y}|}$ and softmax probabilities as $p(\psi(x)) = \operatorname{Softmax}(f(\psi(x))) \in [0, 1]^{|\mathcal{Y}|}$. For simplicity, we will use $p(\psi(x))$ and $p(x)$ interchangeably. We also denote indices for the last two dimensions in tensors as superscripts (e.g., last two dimensions of sizes $\text{ H}, \text{ W}$ for $x$) and all other dimensions as subscripts (e.g., first dimension of size $3$ in $x$). We will use shorthands ``FG'' and ``BG'' for foreground and background, respectively.

\subsection{Method}
\label{sec:method:method}

Our Object-Centric Classification with Applied Masks (\ourmethodname{}) pipeline is summarized in Figure~\ref{fig:method_figure}. We use object-centric representations to reduce spurious correlations in image classification.  
It consists of two main parts: 1. generate object-centric representations, 2. perform robust classification by classifying an image using only representations of the foreground object. In the following subsections, we will explain these parts in more detail.

\begin{table*}[t]
\centering
\small
\setlength{\tabcolsep}{1em}
\begin{tabular}{l|llc|c|cccc}\toprule
\textbf{Methods}&   \multicolumn{2}{l}{\textbf{Pre-training Datasets}}&&\textbf{FT}&\multicolumn{2}{c}{\textbf{Movi-C}}& \multicolumn{2}{c}{\textbf{Movi-E}}\\
        &   Encoder& Decoder&&&FG-ARI  &mBO & FG-ARI  &mBO \\
\midrule
Slot Diffusion \citep{SlotDiffusion} &   OpenImages (1.9M)& COCO (118k)&&\cross&66.9  &43.6 & 67.6&26.4\\
Dinosaur \citep{Dinosaur}    &   GLD (1.2M)& COCO (118k)&&\cross&67.0  &34.5 & 71.1&24.2\\
FT-Dinosaur \citep{ft-dinosaur}   &   GLD (1.2M)& COCO (118k)&&\tick&73.3  &44.2 & 71.1&29.9\\
\midrule
HQES \citep{EntitySeg} (Ours)    &   \multicolumn{2}{l}{ COCO (118k) + EntitySeg (33k)}&&\cross&79.3  &65.4 & \textbf{87.2}&63.8\\ 
SAM \citep{SAM} &   \multicolumn{2}{l}{SA-1b (11M) }&&\cross&\textbf{79.7} & \textbf{73.5} & 84.7&\textbf{69.7}\\ \bottomrule
 \end{tabular}
\caption{\textbf{Object Discovery Performance.} Quantitative results for object discovery on Movi-C and Movi-E; column ``FT'' indicates whether the model was fine-tuned on the training split of the corresponding dataset (Movi-C or Movi-E). HQES outperforms the OCL baselines like Slot Diffusion and Dinosaur, despite being sample-efficient (151k training samples).} 
\label{tab:quant_object_discovery}
\end{table*}

\subsubsection{Generating object-centric representations}
\label{sec:method:generate_representations}

To generate the object-centric representations, we first generate masks for all objects and backgrounds in the image using a mask generator. We then apply generated masks to images by combining masks with images. Each object is then encoded with an image encoder.\vspace{0.1cm}

\noindent\textbf{Generating masks.} To produce object representations given an original image 
$x \in [0, 1]^{[3, \text{ H}, \text{ W}]}$
, we generate a set of masks for all the foreground objects and the background.
That is done with the help of a mask generator $S$, which takes $x$ as input and assigns each pixel in $x$ to one of \maxobj{} masks.
The output of this model is the stack of $K$ binary masks, with each mask $m$ corresponding to a different object: $m \in \{S_i, i=1\ldots \numobj\}, m \in \{0, 1\}^{[H,\ W]}$.
An OCL method like FT-Dinosaur \citep{ft-dinosaur} or an external segmentation model like High-Quality Entity Segmentation (HQES)
\citep{EntitySeg} can be used as a mask generator in this pipeline. We will call the mask generator as the mask model or the masking method interchangeably. \vspace{0.1cm}

\noindent\textbf{Applying masks.}
After producing the binary masks for each object, we segregate the pixel contents for each mask by applying the mask on the input image. We will interchangeably call the mask applying operation as the mask method throughout the paper. One way to apply masks to images is to simply add a gray background to all but selected pixels, cropping the image that follows the mask contours, and resizing the result to the size of the original image. In such a case, we call the operation \cropresize.\vspace{0.1cm}

\noindent However, a mask method can be any operation involving an image $x$ and a mask $m$:
$a(x, m) \in [0,1]^{[3,\ H,\ W]}$.
We additionally show ease-of-use in incorporating the latest masking techniques like AlphaCLIP, which combines a mask and original image by appending masks as an additional $\alpha$-channel to the image tensor, resulting in an RGB-A 4-dimensional tensor. This allows using masks as a source of focus instead of removing backgrounds entirely, useful for some practical applications. We call such an operation as ``$\alpha$-channel".\vspace{0.1cm}

\noindent\textbf{Encoding applied masks.}
To get the final object-centric representations we encode applied masks by an image encoder $\psi$ such as ViT \citep{ViT} for example.

\subsubsection{Robust classifier}
\label{sec:method:robust_classifier}

We hypothesize that by isolating foreground object representations from the representations of background and other objects, we eliminate sources of spurious correlations, hence performing more robust classification. For that reason, we first use the set of object-centric representations obtained in the previous stage to select the single representation that corresponds to the foreground. Then we provide the selected foreground representation to the classifier to make the final prediction.\vspace{0.1cm}

\noindent\textbf{FG detector.}
After applying masks to the image, we select the mask that corresponds to the foreground object by the following process. At first, we compute the \textit{foreground score} 
that reflects how likely a given applied mask is to correspond to the foreground object. Then we take the mask with the highest foreground score among all masks for the current image and use it for robust classification.\vspace{0.1cm}

\noindent
Currently, we use two types of foreground scores, both computed from the classifier's outputs:

\begin{enumerate}
    \item \textbf{\highlightpurple{Ens. $\mathcal{H}$}}: 
    $g_{\mathcal{H}}(x, m) = 
    \frac{1}{M} \sum_{k=1}^M \mathcal{H}[p_k(\psi(a(x, m)))]$ - ensemble entropy (see details in §~\ref{sec:ood_detection_main}). Here, \(M\) is the ensemble size, and \(\mathcal{H}\) stands for entropy.
     \item \textbf{\highlightblue{Class-Aided}}: $g_{\text{class\_aided}}(x, m) = p^y(\psi(a(x, m)))$ - probability of predicting a ground truth label. We consider this foreground score to measure the efficacy of the object-centric representation rather than to suggest it as a final method to use in practice. Although in reality, we do not have access to ground truth labels, it provides critical signals as to whether the insufficient generalization performance is due to object representation or due to foreground selection and the classifier.
    
\end{enumerate}

\noindent For the comparison of different foreground scores, see §~\ref{sec:ood_detection_main}.

\begin{figure*}[t]
    \newcommand{\myig}[1]{\adjustbox{valign=c}{\includegraphics[width=0.15\textwidth]{figures/images_movie/#1}}}
    \newcommand{\mycaption}[1]{{\footnotesize#1}}
    \newcommand{\DINOSAUR}{\textsc{Dinosaur}\xspace}
    \newcommand{\SPOT}{\textsc{Spot}\xspace}
    \newcommand{\SAM}{\textsc{Sam}\xspace}
    \newcommand{\SAMcomp}{\textsc{Sam} \textit{(comp.)}\xspace}
    \newcommand{\SAMbest}{\textsc{Sam} \textit{(best.)}\xspace}
    \newcommand{\DINO}{\textsc{Dino}\xspace}
    \newcommand{\COCO}{\textsc{Coco}\xspace}
    \newcommand{\PASCALVOC}{\textsc{Pascal Voc}\xspace}
    \newcommand{\ScanNet}{\textsc{ScanNet}\xspace}
    \newcommand{\ClevrTex}{\textsc{ClevrTex}\xspace}
    \newcommand{\EntitySeg}{\textsc{EntitySeg}\xspace}
    \newcommand{\YCB}{\textsc{Ycb}\xspace}
    \newcommand{\MOVi}{\textsc{Mov}i\xspace}
    \newcommand{\methodshort}{FT-\textsc{Dinosaur}\xspace}

    \renewcommand{\arraystretch}{4.7}
    \setlength{\tabcolsep}{2pt}
    \centering
    \vspace{-0.1\textwidth}
    \begin{tabular}{@{}cccccc@{}}
                \mycaption{Image} & \mycaption{\DINOSAUR{}} & \mycaption{SlotDiffusion} & \mycaption{\methodshort{}}  & \mycaption{\SAM} & \mycaption{HQES} \\[-1.5em]
    
      \myig{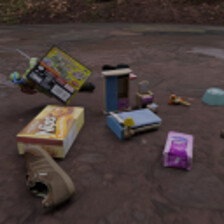} &
      \myig{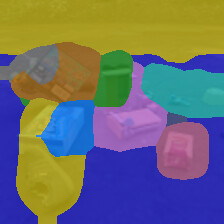} &
      \myig{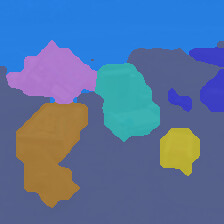} &
      \myig{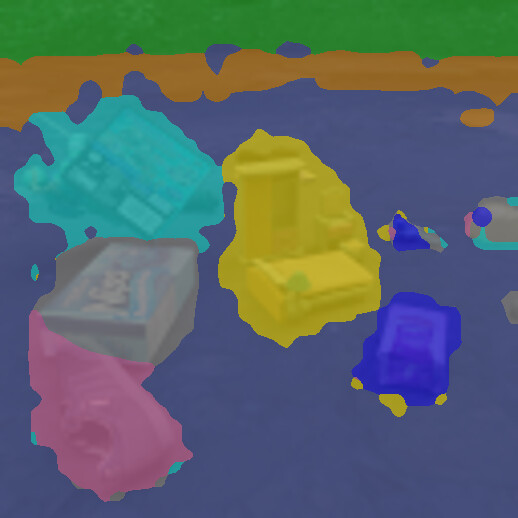} &
     \myig{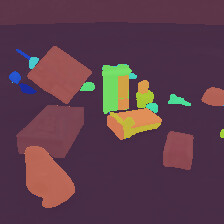} &
      \myig{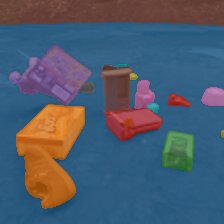} 
      \vspace{.3em}
      \\
    
      \myig{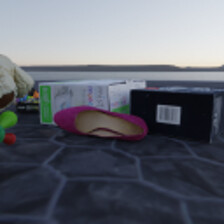} &
      \myig{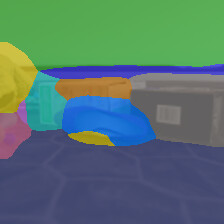} &
      \myig{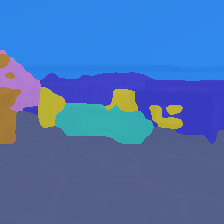} &
      \myig{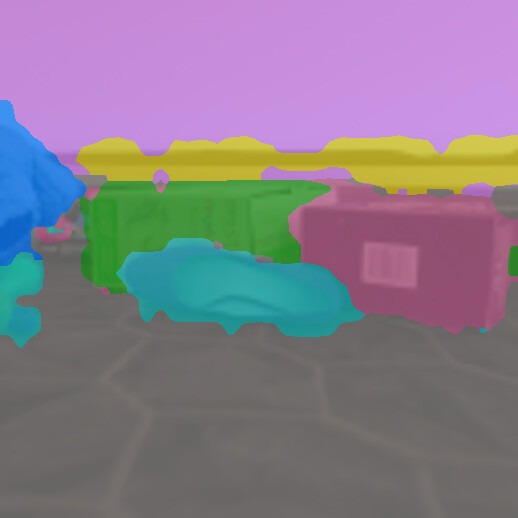} &
      \myig{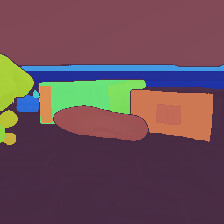} &
      \myig{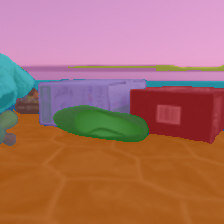} 
      \vspace{.3em}
      \\
    
      \myig{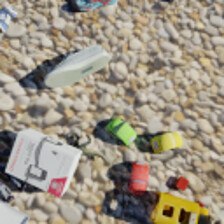} &
      \myig{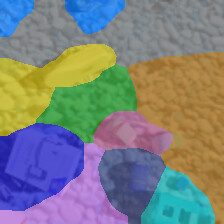} &
      \myig{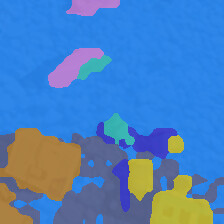} &
      \myig{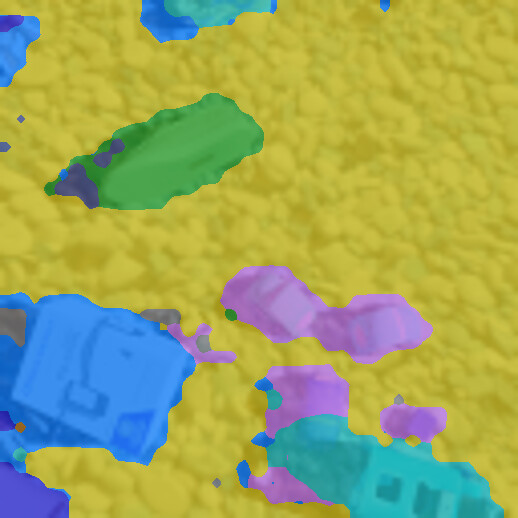} &
      \myig{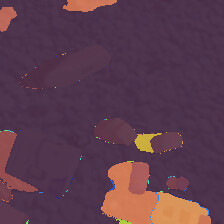} &
      \myig{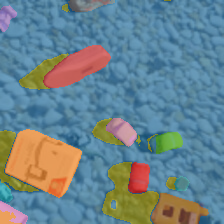} 
      \vspace{.3em}
      \\
    
      \myig{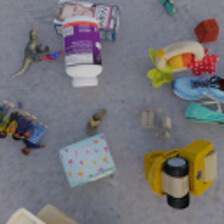} &
      \myig{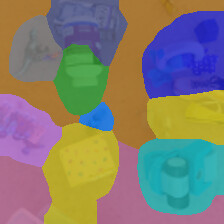} &
      \myig{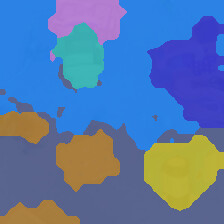} &
      \myig{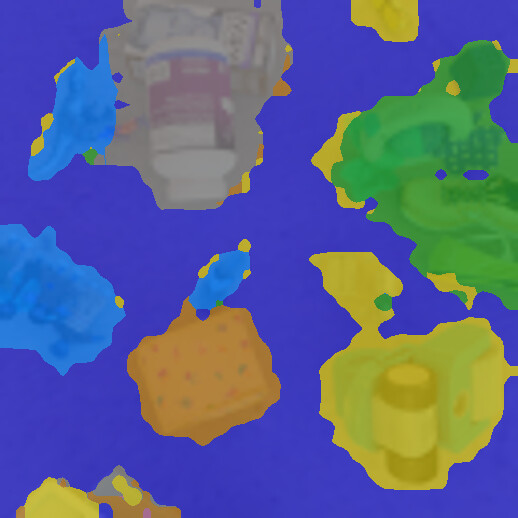} &
      \myig{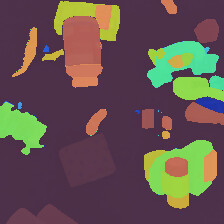} &
      \myig{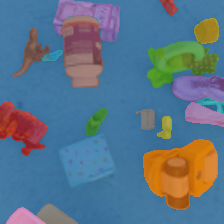}
      \vspace{.3em}
      \\
    
    \end{tabular}
    \caption{\textbf{Qualitative Results on Object Discovery}. 
    \mycaption{\DINOSAUR{}}, \mycaption{SlotDiffusion}, and \mycaption{\methodshort{}} are existing object-centric learning (OCL) approaches. \mycaption{\SAM} and \mycaption{HQES} refer to zero-shot segmentation methods. Images are from \MOVi-E. \mycaption{\SAM} and HQES masks fit objects much better than the masks predicted by OCL methods. All columns except for HQES are taken from \citep{ft-dinosaur}.}
    \label{tab:segmentation_qualitative_movi_e}
\end{figure*}

\vspace{0.5em}
\noindent\textbf{Image classification using FG object representations.}
Finally, once we have identified the mask that matches the foreground object, we apply it to the original image and classify the result of this operation.
The final output of our method is:
$$\text{\ourmethodname{}} (x) = p(\psi(a(x, m^\star)), $$
where $m^\star$ is the mask selected by the FG detector.

%% file: sec/4_experiments.tex
\section{Experiments}

In this section, we first evaluate slot-centric OCL approaches and foundational segmentation models on unsupervised object discovery tasks. We then evaluate whether OCL methods provide robust object classification by benchmarking them against a strong baseline that uses mask predictions from foundational segmentation models, following the \ourmethodname{} pipeline (§\ref{sec:method}). 

\subsection{Are we done with object-discovery?}
\label{sec:object_discovery}

OCL methods are often evaluated by how well they perform on unsupervised object discovery, measured via instance segmentation for every object in the scene. We explore whether the emergence of strong zero-shot segmentation models (class-agnostic) such as HQES \citep{EntitySeg} and SAM \citep{SAM} allows reliable decomposition of the scene into objects. We compare these foundational segmenters against state-of-the-art OCL approaches \citep{SlotDiffusion, Dinosaur, ft-dinosaur}.\vspace{0.1cm}

\noindent\textbf{Setup.} We first describe our experimental setup, including datasets, metrics, and compared baselines. Following prior work \citep{SAVI, SAVI++, Dinosaur, ft-dinosaur}, we use two synthetic image datasets from \citep{Kubric}: Movi-C and Movi-E. Both feature around 1,000 realistic 3D-scanned objects placed on high-definition backgrounds. Movi-C contains 3 – 10 objects per scene, while Movi-E contains 11 – 23.
We quantify model performance using two standard metrics (Table~\ref{tab:quant_object_discovery}): the foreground adjusted Rand index (FG-ARI) \citep{FG_ARI_1971, FG_ARI_1985, SAVI} and mean best overlap (mBO) \citep{mbo_metric, Dinosaur}, detailed in Section §\ref{sec:related_work}. Unlike FG-ARI, mBO also accounts for background pixels. It also measures how well masks fit objects. We compare HQES and SAM to state-of-the-art OCL methods with demonstrated real-world applicability: SlotDiffusion \citep{SlotDiffusion}, Dinosaur \citep{Dinosaur}, and FT-Dinosaur \citep{ft-dinosaur}, all described in Section §\ref{sec:related_work}.\vspace{0.1cm}

\noindent\textbf{Results.} Table \ref{tab:quant_object_discovery} and Figure~\ref{tab:segmentation_qualitative_movi_e} show quantitative and qualitative results. Across both metrics, FG-ARI and mBO across out-of-distribution benchmarks like Movi-C and Movi-E, HQES far surpasses the OCL baselines.  This gap is especially notable in mBO on Movi-E, improving 29.9\% to 63.8\%. Qualitatively, HQES masks fit objects much better than masks predicted by OCL methods (Figure~\ref{tab:segmentation_qualitative_movi_e}). HQES also shows it is possible to be sample efficient, only being trained on 151k samples in contrast to 11M samples for SAM. \vspace{0.1cm}

\noindent\textbf{Conclusion.} Sample-efficient segmentation models, even in a zero-shot setting, excel at object discovery, surpassing OCL methods by large margins. This suggests that one key aspect of OCL — decomposing the scene into objects — can be largely solved by powerful pre-trained segmentation models, effectively replacing the slot-based OCL methods. Given the decomposition, we explore in the next section downstream applications where OCL methods can contribute a lot of practical value.

\begin{table*}[ht]
\vspace{-1em}
\centering
\small
\centering
\begin{tabular}{lc}
\multicolumn{2}{c}{\textbf{(a) ImgNet-D (BG) \citep{IN-D}}} \\
\toprule
   Method & Acc. ($\uparrow$) \\
   \midrule
 \multicolumn{2}{c}{CLIP ViT-L} \\
 \midrule
 CLIP \citep{CLIP} & 23.5\\
 O-D (Ours) & 57.7\\
 O-H (Ours) & 68.0\\
 CLIP-SigLip \citep{SigLip} & 59.4\\
 O-D-SigLip (Ours) & 71.5\\
 O-H-SigLip (Ours) & \textbf{78.5}\\
 \midrule
 \multicolumn{2}{c}{Multi-modal LLMs} \\
 \midrule
 MiniGPT-4 \citep{minigpt4} & 71.8 \\
 LLaVa \citep{LLAVA} & 52.9 \\
 LLaVa-NeXT \citep{LLAVA_NEXT} & 68.8 \\
 LLaVa-1.5 \citep{LLAVA1_5} & 73.3$^{\star}$ \\

 \bottomrule
\end{tabular}
\hspace{0em}
\hfill
\centering
\begin{tabular}{lc}
\multicolumn{2}{c}{\textbf{(b) UrbanCars \citep{Whaca}}} \\
\toprule
 Method & WGA ($\uparrow$) \\
 \midrule
 \multicolumn{2}{c}{ViT-L-14 CLIP} \\
 \midrule
 CLIP \citep{CLIP} & 87.2\\
 O-D (Ours) & 98.4\\
 O-H (Ours) & \textbf{100.0}\\
 \midrule
 \multicolumn{2}{c}{ResNet50 CLIP} \\
 \midrule
 CLIP \citep{CLIP} & 64.8\\
 O-D (Ours) & 98.4\\
 O-H (Ours) & \textbf{100.0}\\
 \midrule
 \multicolumn{2}{c}{ResNet50} \\
 \midrule
 CoBalT \citep{CoBalT} & 80.0 \\
 LfF \citep{LfF2020} & 34.0 \\
 JTT \citep{JTT} & 55.8 \\
 SPARE \citep{SPARE} & 76.9 \\
 LLE \citep{Whaca} & 90.8$^{\star}$ \\
 \bottomrule
\end{tabular}
\hspace{0em}
\hfill
\centering
\begin{tabular}{lc}
\multicolumn{2}{c}{\textbf{(c) ImgNet-9 (MR) \citep{IN-9}}} \\
\toprule
 Method & Acc. ($\uparrow$) \\
 \midrule
 \multicolumn{2}{c}{ViT-L-14 CLIP} \\
 \midrule
 CLIP \citep{CLIP} & 91.9\\
 O-D (Ours) & 93.8\\
 O-H (Ours) & \textbf{95.2}\\
 \midrule
 \multicolumn{2}{c}{ResNet50 CLIP} \\
 \midrule
 CLIP \citep{CLIP} & 81.1\\
 O-D (Ours) & 80.6\\
 O-H (Ours) & \textbf{85.6}\\
 \midrule
 \multicolumn{2}{c}{ResNet50} \\
 \midrule
 CoBalT \citep{CoBalT} & 80.3 \\
 SIN \citep{SIN} & 63.7 \\
 INSIN \citep{SIN} & 78.5 \\
 INCGN \citep{SIN} & 80.1 \\
 MaskTune \citep{MASKTUNE} & 78.6 \\
 CIM \citep{CIM} & 81.1$^{\star}$ \\
 \bottomrule
\end{tabular}
\hspace{0em}
\hfill
\centering
\begin{tabular}{lc}
\multicolumn{2}{c}{\textbf{(d) Waterbirds \citep{Waterbirds}}} \\
\toprule
 Method& WGA ($\uparrow$) \\
 \midrule
 \multicolumn{2}{c}{ViT-L-14 CLIP}\\
\midrule
  CLIP \citep{CLIP} & 83.6\\
  O-D (Ours)&92.1\\ 
  O-H (Ours)& \textbf{96.0}\\
  \midrule
 \multicolumn{2}{c}{ResNet50 CLIP}\\
 \midrule
 CLIP \citep{CLIP}&72.9\\
 O-D (Ours)&83.3\\
 O-H (Ours)&\textbf{92.5}\\
  \midrule
 \multicolumn{2}{c}{ResNet50}\\
  \midrule
  CoBalT \citep{CoBalT}& 90.6\\
 GDRO \citep{Waterbirds}&89.9\\
 AFR \citep{AFR} &90.4\\
 SPARE \citep{SPARE} &89.8\\
 MaskTune \citep{MASKTUNE} &86.4\\
 CIM \citep{CIM}&77.2\\
 DFR \citep{DFR} &91.8$^{\star}$\\ 
 \bottomrule
\end{tabular}
\vspace{-.5em}

\caption{\textbf{Object-Centric Learning for Spurious Background OOD Generalization}. 
We report versions of accuracy in each benchmark. Results are grouped according to backbone architecture. 
``ImgNet-D (BG)'' stands for the ImageNet-D ``background'' subset. ``ImgNet-9 (MR)'' stands for the ImageNet-9 ``mixed rand'' subset. ``WGA'' stands for the worst group accuracies. 
O-H/O-D stands for OCCAM with HQES/FT-Dinosaur masks generator correspondingly.
For cited methods, we show results reported in the papers \citep{CoBalT} and \citep{IN-D}. $\star$ indicates the state-of-the-art results in each benchmark.}
\label{tab:all_results}
\end{table*}

\subsection{Application: Classification with Spurious Background Correlations}
\label{subsec:application-spurious-background}

As foundational segmentation models outperform OCL methods in decomposing the scene into constituent objects, we take a further step and evaluate OCL methods on a downstream task that leverages the disentangled representations for distinct objects: robust classification under spurious background cues. This subsection demonstrates that object masks are a simple but effective strategy to  mitigate the influence of spurious correlations with backgrounds in classification tasks (Table~\ref{tab:all_results}).\vspace{0.1cm}

\noindent\textbf{Setup.} We first describe our experimental setup, including datasets, metrics, and compared baselines. We use several standard datasets with spurious backgrounds or co-occurring objects — UrbanCars \citep{Whaca}, ImageNet-D (background subset) \citep{IN-D}, ImageNet-9 (mixed rand subset) \citep{IN-9}, Waterbirds \citep{Waterbirds}, and CounterAnimals \citep{CounterAnimals} — detailed further in §\ref{sec:ext_spurious_bg}. We measure model performance using the standard metric used in the respective benchmark: accuracy and worst group accuracy (WGA). We provide per-benchmark comparisons for reference, including results from other relevant methods, citing them alongside their names in the tables. 
We use the foundational segmentation model HQES \citep{EntitySeg} (O-H) and the state-of-the-art OCL method FT-Dinosaur \citep{ft-dinosaur} (O-D) for mask prediction in our training-free probe, OCCAM. We categorize methods with comparable image encoder backbones for fairness.\vspace{0.1cm}

\noindent\textbf{Results.} Using masks significantly improves performance across all datasets, sometimes reaching 100\% accuracy (e.g., on UrbanCars; Table~\ref{tab:all_results}(b)) or close to that performance on Waterbirds and ImageNet-9 (mixed rand) subsets. This shows the potential of simple, training-free object-centric methods like OCCAM to address otherwise challenging downstream problems, if we can robustly identify the foreground object of interest. On harder benchmarks like ImageNet-D (background subset), HQES-based masks with SigLip models yield far better performance (78.5\%) even compared to recent models like LLAVA 1.5 \citep{LLAVA} (73.3\%), and outperform their best slot-based counterparts (71.5\%) 
using FT-Dinosaur (Table~\ref{tab:all_results}(a)). Throughout, HQES consistently provides more effective masks than FT-Dinosaur.\vspace{0.1cm}

\noindent\textbf{Conclusion.} These experiments show that mask-based, training-free object-centric probes can provide practical value on challenging robust classification tasks, if the task of foreground detection is sufficiently addressed (§\ref{sec:method:robust_classifier}). It provides substantial gains on all tested benchmarks over the state-of-the-art methods for tackling spurious correlations. We hope this encourages the community to develop segmentation-based OCL approaches and demonstrate practical benefits across a variety of downstream applications. We next perform data-centric analysis leveraging properties of our OCL pipeline.

\subsubsection{CounterAnimals: Spurious or Simply Hard?}
\label{sec:exp:counter_animal}

Our object-centric classification pipeline can isolate an object’s influence apart from its background. This property of OCL can be used to analyze the recently proposed CounterAnimals dataset \citep{CounterAnimals}.\vspace{0.1cm}

\begin{table}
\vspace{-0.5cm}
    \centering
    \small
    \begin{tabular}{lcc}
\multicolumn{3}{c}{\textbf{CounterAnimals}} \\
\toprule
 Method& Cmn/Cntr ($\uparrow$) & Cmn-Ctr ($\downarrow$) \\
 \midrule
 \multicolumn{3}{c}{AlphaCLIP ViT-L} \\ \midrule
  CLIP \citep{CLIP} & 79.0/62.0& 17.0\\
  O-D (Ours)& \textbf{85.8}/\textbf{70.5}& 15.3\\ 
  O-H (Ours)& 84.4/69.2& 15.2\\
 \bottomrule
\end{tabular}
    \caption{\textbf{Data-Centric Understanding using OCL.} We report the accuracies on the \highlightgreen{Common} and \highlightred{Counter} subset of the CounterAnimals dataset. We see that after eliminating the spurious background using OCL methods, the gap (Cmn-Ctr) does not substantially decrease.}
    \label{tab:counteranimals}
\end{table}
\noindent\textbf{Setup.} CounterAnimals highlights models’ reliance on spurious backgrounds. It consists of two splits from iNaturalist,\footnote{\href{https://www.inaturalist.org/observations}{https://www.inaturalist.org/observations}} each containing animals from 45 classes in ImageNet-1k \citep{ImageNet}. The \highlightgreen{Common} split features typical backgrounds (e.g., polar bears on snow), while the \highlightred{Counter} split features less common ones (e.g., polar bears on dirt). It primarily demonstrates that models consistently perform better on the \highlightgreen{Common} than on \highlightred{Counter}, due to spurious background cues.\vspace{0.1cm}

\begin{table*}[t]
\centering
\small
\begin{tabular}{l|ll|l|ccccc}
\toprule
\textbf{Name} & \textbf{Mask Method} & \textbf{Mask Model}  & \textbf{FG Detector} & \textbf{WB}$\uparrow$ & \textbf{IN-9}$\uparrow$ & \textbf{IN-D}$\uparrow$ & \textbf{UC}$\uparrow$ & \textbf{Cmn-Ctr}$\downarrow$ \\
\hline
\multirow{5}{*}{CLIP \cite{CLIP}} & -  & -    & -   & 83.6& 91.9& 17.6& 87.2& 15.0\\\cline{2-9}
&  \multirow{4}{*}{Gray BG + Crop} & \multirow{2}{*}{FT-Dinosaur} & Ens. $\mathcal{H}$& 83.8& 84.0& 52.4& 95.2& 13.1\\
            &      &   & Class-Aided  & 92.1& 93.8& 57.7& 98.4& 12.7\\\cline{3-9}
& & \multirow{2}{*}{HQES} & Ens. $\mathcal{H}$& 86.8& 88.6& 60.4& 95.2& 8.8\\
            &      &   & Class-Aided  & 96.0& 95.2& 68.0& 100.0& 8.5\\
 \hline
\multirow{5}{*}{AlphaCLIP \cite{AlphaClip}}  & - ($\alpha=1$)  & -   & - & 79.8& 90.2& 23.5& 87.2& 17.0\\ \cline{2-9}
& \multirow{4}{*}{$\alpha$-channel} & \multirow{2}{*}{FT-Dinosaur}  & Ens. $\mathcal{H}$& 81.0& 90.3& 40.7& 92.0& 17.2\\
           &      &   & Class-Aided  & 86.9& 93.1& 49.1& 96.0& 15.3\\ \cline{3-9}
 &  &  \multirow{2}{*}{HQES}  & Ens. $\mathcal{H}$& 84.7& 91.2& 44.7& 91.2& 16.4\\
            &      &   & Class-Aided                                         & 89.1& 93.1& 53.9& 97.6& 15.2\\
\bottomrule
\end{tabular}
\caption{\textbf{Factor Analysis for Spurious Background OOD Generalization}. Accuracies on spurious correlations datasets when varying factors for the ViT-L-14 CLIP architecture. We use AlphaCLIP for $\alpha$-channel masking and CLIP for Gray Crop masking. We first report their baseline performances without masking (where mask method and model are both ``-'') and with 2 different mask models (FT-Dinosaur and HQES) as well as 2 different foreground detectors (\highlightpurple{Ens. $\mathcal{H}$} and \highlightblue{Class-Aided}). Results are reported on 5 benchmark datasets, Waterbirds (WB), ImageNet-9 (IN-9), ImageNet-D (IN-D), UrbanCars (UC), and CounterAnimals (Cmn-Ctr). For the CounterAnimals results, we report the gap between the common-split (Cmn) and the counter-split (Ctr) accuracies. Unlike other metrics, a smaller Cmn-Ctr gap is deemed a better generalization.}
\label{tab:spurious_correlations_factors}
\end{table*}

\noindent\textbf{What is the Contribution of Spurious Correlations?} We perform a simple check using OCCAM -- If the drop from \highlightgreen{Common} to \highlightred{Counter} is caused by spurious background correlations, then using OCCAM we can ablate the contribution of everything except the foreground object. Ideally, ablating the background should result in roughly equal performance on both \highlightgreen{Common} and \highlightred{Counter} sets (the gap should be 0\%). However, we see from Table \ref{tab:counteranimals}, Table \ref{tab:spurious_correlations_factors} and Figure \ref{fig:counter_animals_gaps} that even after ablating the background entirely, there is a substantial gap between the \highlightgreen{Common} and \highlightred{Counter} subsets. For example, when using AlphaCLIP, the gap reduces from 17.0\% to 15.2\%. Similarly, using HQES masks and a gray background for both sets, we still observe an 8.5\% gap.
This provides interesting evidence that images in the \highlightgreen{Common} subset might be substantially easier than images from the \highlightred{Counter} subset by about 8-10\%. \vspace{0.1cm}

\noindent\textbf{Conclusion.} OCL methods allow analyzing datasets,
and analyse the contribution of individual objects. In the case of CounterAnimals, we find that spurious backgrounds might not be the primary reason the \highlightred{Counter} subset is harder, although they are a factor. A significant (10\%) gap might be caused by the \highlightred{Counter} subset simply being harder to classify than the \highlightgreen{Common} subset due to a wide variety of other factors. Overall, we show the potential for OCL methods to help inform data-centric fields like data attribution.

\subsubsection{Ablations: Identifying Bottlenecks in OCCAM}
\label{sec:exp:mitigation:factors_analysis}

We now ablate the contributions of different components in the OCCAM pipeline. We first test two CLIP models (CLIP and AlphaCLIP), to see whether our results generalize beyond simply removing backgrounds to recent techniques such as AlphaCLIP, which use the $\alpha$-channel to focus on the mask instead of eliminating the background. Secondly, we study the effect of the masking generator, testing HQES along with the current SOTA OCL method, FT-Dinosaur. Lastly, we study the influence of different FG Detection methods. We showcase our analysis in Table~\ref{tab:spurious_correlations_factors}.\vspace{0.1cm} 

\noindent\textbf{Effect of mask applying method.}
Using masks with \highlightblue{Class-Aided} FG detector improves performance on all the datasets for both Gray BG + Crop and $\alpha$-channel mask methods, but for the former, accuracy is usually higher. For example, on Waterbirds (Table~\ref{tab:spurious_correlations_factors}), accuracy for the Gray BG + Crop mask method and the HQES mask generator is $96.0\%$ while for AlphaCLIP it is $89.1\%$. This indicates that the backgrounds have strong spurious correlations that still affect $\alpha$-CLIP to a small extent. \vspace{0.1cm} 

\noindent \textbf{Effect of mask generator.} Comparing the rows from mask models to the original CLIP model, we see that both FT-Dinosaur and HQES improve performance, across CLIP and AlphaCLIP, given that we use \highlightblue{Class-Aided} FG detector. In this scenario, HQES improves accuracy more than FT-Dinosaur. For example, for the Gray BG + Crop mask method, it leads to $68.0\%$ accuracy on ImageNet-D, while FT-Dinosaur reaches only $57.7\%$. This indicates that the segmentation-based OCL performs better consistently for downstream OCL applications. \vspace{0.1cm} 

\noindent\textbf{Selecting foreground mask.}
Accuracy gains with \highlightpurple{Ens. $\mathcal{H}$} are always smaller than for \highlightblue{Class-Aided} FG detector and sometimes can be negative (Table~\ref{tab:spurious_correlations_factors}). For example, for Gray BG + Crop mask method and HQES mask generator accuracy on ImageNet-9 drops from $91.9\%$ to $88.6\%$ when using \highlightpurple{Ens. $\mathcal{H}$} FG detector, while jumping to $95.2\%$ with \highlightblue{Class-Aided} FG detector. Such results are not surprising at all, given that HQES with \highlightblue{Class-Aided} foreground detector is a very close approximation to classifying ground truth foreground objects (see §~\ref{sec:non_fg} for details). At the same time, this reveals a weakness in other baseline foreground detection methods and leaves room for improvement and future research.
\vspace{0.1cm} 

\noindent \textbf{Conclusion.}
The empirical results show that segmentation models outperform current OCL methods in obtaining object-centric representations that result in better classification. The simple Gray BG + Crop mask method generally performs better than the more advanced $\alpha$-channel mask method.
At the same time, identifying foreground masks among many candidates remains a challenge. 

\subsection{Foreground Detectors Comparison}
\label{sec:ood_detection_main}

\begin{figure}
    \centering
    \includegraphics[width=0.45\textwidth]{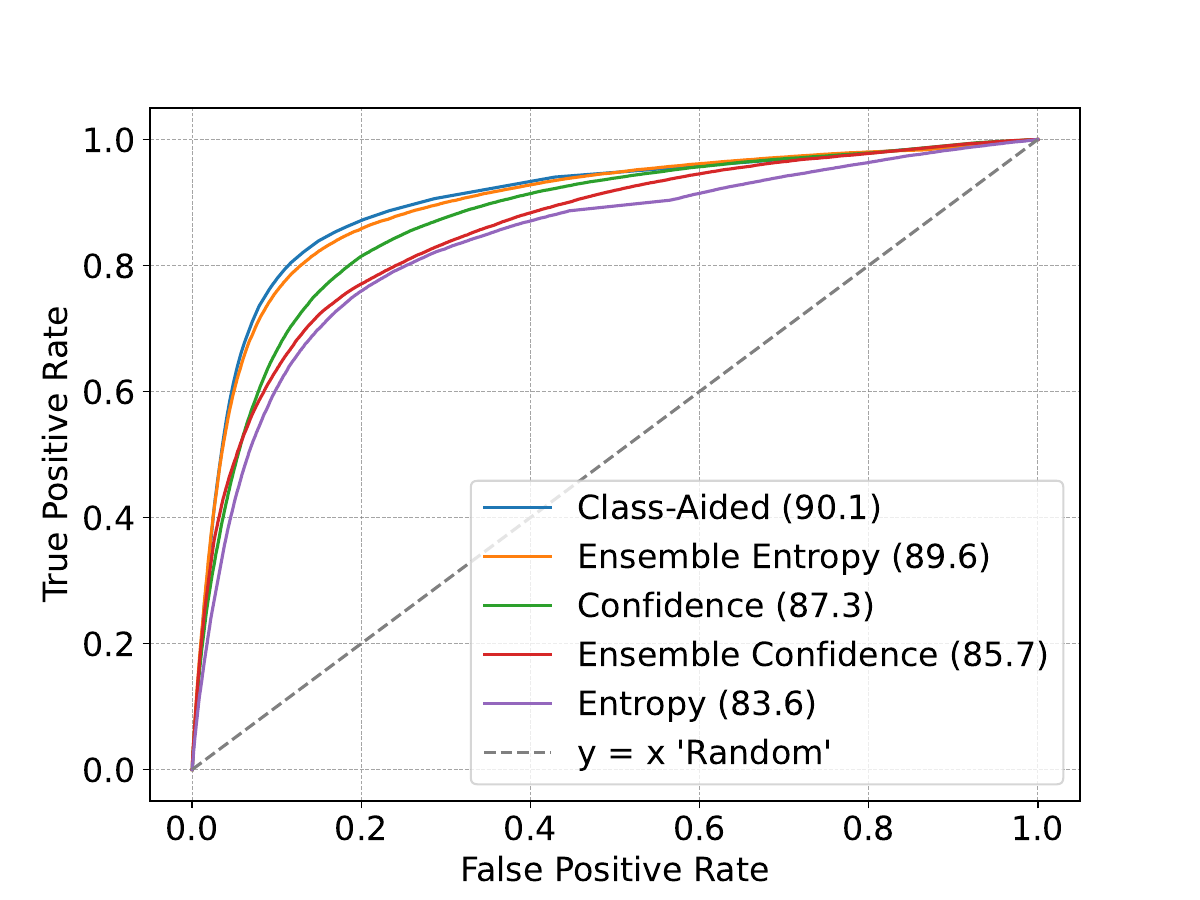}
    \caption{\textbf{Foreground Object Detection.} ROC-curves for foreground detection methods. For each scoring scheme, we measure how well the true foreground objects in the ImageNet-validation dataset are detected. More details in §~\ref{subsec:ood_detection}.}
    \label{fig:foreground_detection}
\end{figure}

To justify the choice of \(g_{\text{class\_aided}}\) and $g_{\mathcal{H}}$ in §~\ref{sec:method:robust_classifier}, we compare several foreground detection methods. One can notice that foreground detection is an application of an out-of-distribution (OOD) detection, a well-studied problem \citep{Mukhoti2021DeepDU, Tran2022PlexTR, Gruber2022UncertaintyEO} — with foreground objects treated as in-distribution (ID) samples and background objects as OOD samples. Hence, we evaluate OOD detection methods for this task in Figure~\ref{fig:foreground_detection}.\vspace{0.1cm}

\noindent \textbf{Setup.} We construct an OOD detection dataset using the ImageNet-1k \citep{ImageNet} validation set by leveraging ground truth bounding boxes\footnote{\href{https://academictorrents.com/details/dfa9ab2528ce76b907047aa8cf8fc792852facb9}{https://academictorrents.com/details/dfa9ab25}} to derive accurate foreground masks (see details in §~\ref{subsec:ood_detection}). Performance is measured via the area under the ROC curve (AUROC), in line with standard OOD detection frameworks \citep{Mukhoti2021DeepDU, Tran2022PlexTR, Gruber2022UncertaintyEO, BalintUncDecomp, rubinstein2024scalable}. We use the following strong baselines:\vspace{0.1cm}  
\begin{itemize}
    \item \textit{Class-Aided (single model) }\citep{hendrycks2017a}: \(p^y(x)\)
    \item \textit{Ensemble entropy }\citep{ovadia2019can}: \(\frac{1}{M} \sum_{k=1}^M \mathcal{H}[p_k(x)]\)
    \item \textit{Ensemble confidence }\citep{Lakshminarayanan2017}: \(\max_c \frac{1}{M} \sum_{k=1}^M p^c_k(x)\)
    \item \textit{Confidence (single model) }\citep{hendrycks2017a}: \(\max_c\, p^c(x)\)
    \item \textit{Entropy (single model) }\citep{Depeweg2017DecompositionOU}: \(\mathcal{H}[p(x)]\)    
\end{itemize}\vspace{0.1cm}

\noindent Here, \(y\) is ground truth label, \(p(x)\) denotes the model's probability vector prediction for the corresponding sample \(x\), \(M\) is the ensemble size, and \(\mathcal{H}\) represents entropy. We use the ViT-L-14 CLIP model pre-trained by OpenAI \citep{CLIP} as the single model, and 5 CLIP models with ViT-L-14 \citep{ViT} vision encoders pre-trained on different datasets as the ensemble. Note that OpenAI ViT-L-14 was the strongest model by AUROC among the ensemble, hence it was used as the single model. Further details are provided in §~\ref{subsec:ood_detection}. \vspace{0.1cm}

\noindent \textbf{Results.} As shown in Figure~\ref{subsec:ood_detection}, \highlightblue{Class-Aided} achieves the highest AUROC of \(90.1\%\) whereas the ensemble entropy method yields \(89.6\%\). Other methods perform significantly worse. Nevertheless, all methods score more than \(80\%\) AUROC.

\noindent \textbf{Conclusion.} The AUROC performance of \highlightblue{Class-Aided} and \highlightpurple{Ens. $\mathcal{H}$} foreground detectors showed only minor differences from each other, both scoring around \(90\%\) and being the best among the compared methods; however, substantial performance gaps remain when comparing the \highlightblue{Class-Aided} results with the \highlightpurple{Ens. $\mathcal{H}$} foreground detector in spurious correlation tasks (Table~\ref{tab:spurious_correlations_factors}), a possible reason for this is discussed in §~\ref{sec:non_fg}. This disparity highlights two key implications. Current evaluation metrics may have a large research gap to better reflect real-world applications. Conversely, spurious correlation foreground detection might be a promising proxy task for identifying better OOD detection models.

%% file: sec/5_discussions.tex
\section{Discussion}

\noindent \textbf{In defense of current OCL benchmarks.} One important aspect to clarify is the rationale behind the OCL researchers' choice to evaluate their models using object discovery benchmarks, as this may not have been clearly articulated. Conventionally, OCL works have relied on constructing synthetic scenarios, where one has knowledge of the ground truth object-centric latent variables, e.g., object position, object color, etc, and can thus directly evaluate whether the learned representation encodes each object separately in its representation \citep{brady2023provably, kori2023grounded, brady2024interaction}. One core aspect is scaling it to real-world scenarios, where we do not have knowledge of the data-generating process. Hence, traditional literature resorts to (a) probing the representation for object properties, such as object position, object color, etc \citep{CoBalT, liu2023causal}, and (b) decoding slot representations to observe if they do indeed only possess a given object \citeobjectdiscovery. \vspace{0.1cm}

\noindent \textbf{Should OCL be strictly unsupervised?} Traditionally, it was assumed that without access to auxiliary information or data-generating processes, there could be no ground-truth supervision for object-centric representations. Consequently, unsupervised learning — requiring no labels — became the standard approach for OCL. However, the advent of robust foundation models — that can leverage segmentation masks or text alongside images and generalize zero-shot across a wide range of inputs — now challenges the need for strict unsupervised constraints \citep{Ferdinand1}. We believe OCL can greatly benefit from using all available data.\vspace{0.1cm}

\noindent \textbf{Why not incorporate developmentally plausible multimodal cues in OCL?} When modeling human-like object perception, we should focus on developmentally plausible supervision. However, we note that the assumption of visual learning in infants being unsupervised also warrants reconsideration. Infants do not learn solely from static images; rather, they integrate a wealth of sensory cues (see \citet{ayzenberg2024development} for a detailed review). For example, Spelke’s seminal review \citep{spelke1990principles} highlights the importance of dynamic information, such as motion and depth cues, for effective object segmentation in early development. Some object-centric works (e.g. \citet{didolkar2023cycle}) argue against this primarily based on the feasibility,  citing the unavailability of multimodal data. However, there are several computational studies with models incorporating motion or depth (e.g. \citet{karazija2022unsupervised, SAVI++}), which also demonstrate that these additional cues can, in fact, be leveraged effectively. Thus, there is no inherent reason to confine OCL to strictly unsupervised, image-only paradigms when richer, multimodal data is often accessible in practice.

\section{Conclusion and open problems}

The motivation for object-centric learning (OCL) originates from a variety of goals, including out-of-distribution generalization, sample-efficient composition, and insights into human cognitive object perception. Despite this broad scope, progress has been measured mostly by object-discovery benchmarks only. With the advent of strong segmentation methods such as High-Quality Entity Segmentation (HQES) \citep{EntitySeg}, we confirm that class-agnostic segmentation models far surpass slot-based OCL methods in obtaining isolated object representations, effectively meeting OCL’s initial goal.\vspace{0.1cm}

\noindent However, its relevance extends beyond object discovery. We advocate for shifting OCL evaluation towards more realistic downstream tasks that leverage object-centric representations, such as mitigating spurious background correlations. We design a simple training-free probe, \ourmethodname{}, to show the efficacy of object-centric approaches to help classifiers generalise even in the presence of spurious correlations (\S\ref{subsec:application-spurious-background}), achieving near-perfect accuracies across many benchmarks (Table \ref{tab:all_results}). By separating object-wise representation (well-addressed by HQES) from object selection (still a key challenge), \ourmethodname{} sheds light on where further improvements are needed.\vspace{0.1cm}

\noindent Looking ahead, we hope OCL-based approaches benchmark visual understanding through scene-graph construction, more interpretable intermediate representations, and human-in-the-loop feedback for cue selection. We hope diverse applications and creating corresponding benchmarks will push the field forward. Beyond immediate use cases, OCL may also inform fundamental cognitive questions about how objects and causal structures emerge in the real world and how infants understand objects without explicit supervision \citep{CognitionSpelke, InfantsCognition}. Realizing this broader vision will require refining the OCL objective and breaking it down into well-defined subproblems that can further illuminate these deeper inquiries.

\section*{Author Contributions}

Alexander and Ameya conceived the project. Alexander led the experiments, and Joon helped design the experiments. Alexander, Ameya, and Joon led the writing of the paper. Matthias and Joon provided helpful feedback throughout the project.

\section*{Acknowledgments}

The authors would like to thank (in alphabetical order): Michael Kamp, Shyamgopal Karthik, Yash Sharma, Matthias Tangemann, Arnas Uselis, and Thaddaeus Wiedemer for insightful feedback and suggestions. This work was supported by the Tübingen AI Center. AP and MB acknowledge financial support by the Federal Ministry of Education and Research (BMBF), FKZ: 011524085B and Open Philanthropy Foundation, funded by the Good Ventures Foundation. AR thanks the International Max Planck Research School for Intelligent Systems (IMPRS-IS) for support. This research utilized compute resources at the Tübingen Machine Learning Cloud, DFG FKZ INST 37/1057-1 FUGG.

%% file: sec/X_suppl.tex
\clearpage
\onecolumn

\section{CounterAnimals: gaps between ``Common'' and ``Counter'' subsets}
\label{sec:counter_animal}

\begin{figure}
\centering
\includegraphics[width=1.0\linewidth]{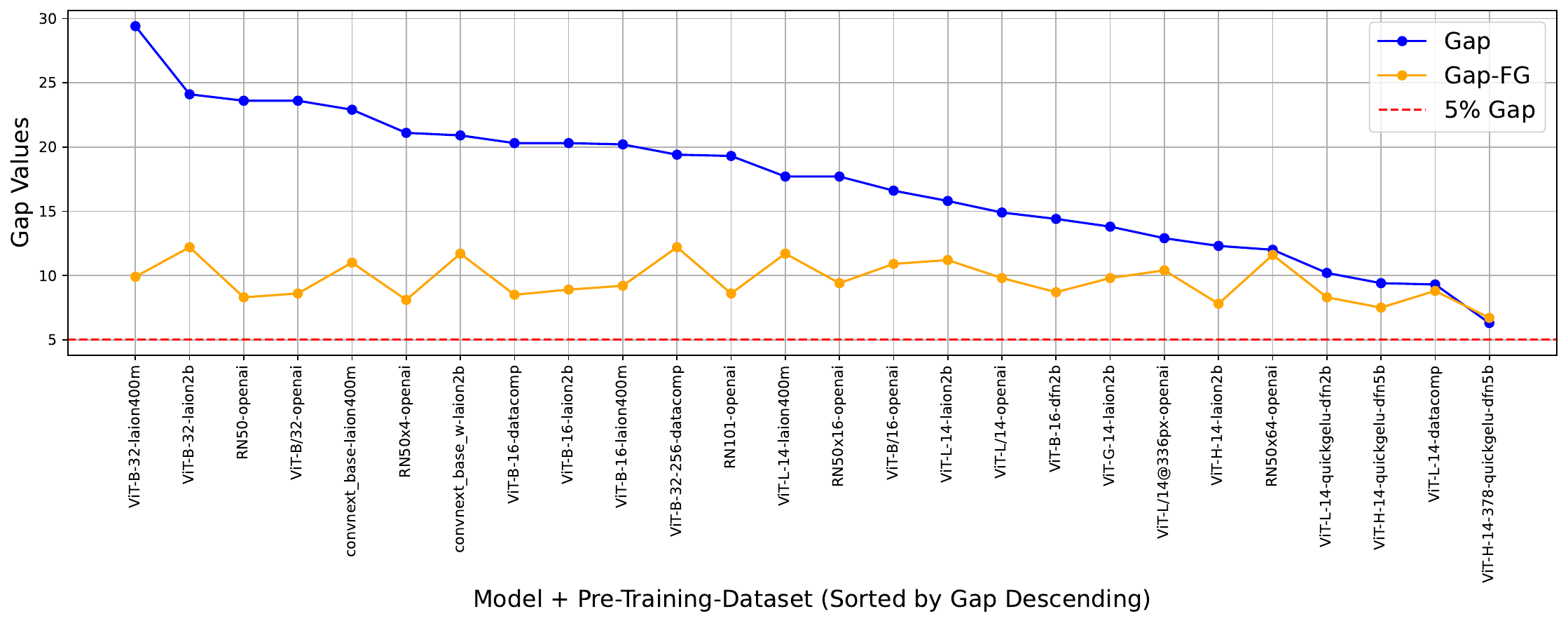}
\caption{Gaps in accuracies [\highlightgreen{Common} - \highlightred{Counter}] for \highlightgreen{Common} and \highlightred{Counter} subsets of CounterAnimals \citep{CounterAnimals} dataset correspondingly for different CLIP models and pre-training datasets. ``Gap'' results are computed for CLIP \citep{CLIP} zero-shot performance without using any masks; ``Gap-FG'' results are computed when using OCCAM with HQES \citep{EntitySeg} masks, \highlightblue{Class-Aided} foreground selection method, and \cropresize{} mask applying operation.}
\label{fig:counter_animals_gaps}
\end{figure}

In addition to the results in Table~\ref{tab:all_results}~(e), we present the complete performance results for all CLIP models from the original CounterAnimals dataset \citep{CounterAnimals} in Figure~\ref{fig:counter_animals_gaps}. This figure illustrates the performance gaps between the \highlightgreen{Common} and \highlightred{Counter} subsets, as discussed in §~\ref{sec:exp:counter_animal}.

\noindent We observe that the performance gaps are consistently greater than $5\%$, as all points lie above the red dashed line. For some models, such as ViT-L-14-datacomp and ViT-H-14-quickgelu-dfn5b, the gaps remain nearly unchanged with or without using OCCAM — around $10\%$ and $6\%$, respectively.

\noindent We argue that for these models, the original gaps reported in the CounterAnimals paper \citep{CounterAnimals} (referred to as “Gap” in our notation) cannot be attributed solely to the models’ reliance on spurious background cues. This is because the gap remains even after background removal using the “Gray BG + crop” masking operation (referred to as “Gap-FG” in our notation).

\clearpage

\section{Details on spurious backgrounds datasets}
\label{sec:ext_spurious_bg}

Below we provide details on the datasets used in our study (for more information on the CounterAnimals dataset, see §~\ref{sec:counter_animal}):

\noindent The core of our dataset collection includes several widely-used benchmarks for evaluating robust image classification models: UrbanCars \citep{Whaca}, Waterbirds \citep{Waterbirds}, and ImageNet-9 \citep{IN-9}. We also include the ImageNet-D dataset \citep{IN-D}, which we consider to offer more realistic visual compositions, as it uses a diffusion model \citep{StableDiffusion} to blend objects with backgrounds, rather than relying on manual cut-and-paste techniques as in the previous datasets. Finally, we use the CounterAnimals dataset \citep{CounterAnimals}, a recently introduced benchmark consisting of natural images with spurious background correlations, specifically designed to challenge even CLIP models.

\begin{enumerate}
    \item \textbf{UrbanCars} \citep{Whaca}: A binary classification dataset that categorizes cars as either ``urban'' or ``country.'' Each image contains a car paired with a contextually related secondary object (e.g., a fire hydrant for urban or a cow for country) and is placed on either an urban or rural background. All elements are synthetically combined from cut-out components.

    \item \textbf{ImageNet-D} \citep{IN-D}: A synthetic dataset generated using diffusion models for 113 ImageNet-based classes (a subset of ImageNet-1k \citep{ImageNet}). We focus on the ``background'' subset, where objects appear in unexpected contexts (e.g., plates in a swimming pool), to test robustness to spurious background cues.

    \item \textbf{ImageNet-9} \citep{IN-9}: A synthetic dataset with 9 broad object categories (e.g., dog, bird), each corresponding to supersets of ImageNet classes. We use the ``mixed random'' subset, where objects are placed on backgrounds from different, unrelated classes.

    \item \textbf{Waterbirds} \citep{Waterbirds}: A binary classification dataset where bird species are labeled as either ``land'' or ``sea'' birds. Each image features a bird placed on either a land or sea background. Like UrbanCars, this dataset is synthetically constructed using cut-out birds and backgrounds.
\end{enumerate}

\clearpage

\section{Class-Aided foreground detector yields the closest approximation to ground truth foreground masks}
\label{sec:non_fg}

\noindent The \highlightblue{Class-Aided} foreground detector selects masks based on the highest ground truth class probability (§~\ref{sec:method:robust_classifier}). 

\noindent Such a strategy may introduce a selection bias towards non-foreground masks that boost the overall classification accuracy of \ourmethodname{}, but are unrelated to the actual objects of interest — for example, masks highlighting spurious background regions that correlate with the ground truth label. For this reason, we were initially cautious about treating it as a reliable foreground detector.

\noindent However, on the Waterbirds dataset \citep{Waterbirds}, for which ground truth foreground masks are available \citep{DFR}, we find that this bias is infrequent. In a random sample of 100 images, \highlightblue{Class-Aided} selected a non-foreground mask in only 5 cases. Despite this, the classification accuracy using \highlightblue{Class-Aided} masks is $96.0\%$, only slightly lower than the $96.7\%$ achieved with ground truth masks (see Table~\ref{tab:non_fg}). 

\noindent Based on this, we do not observe strong evidence that the \highlightblue{Class-Aided} detector frequently selects non-foreground masks, whereas we find that the selected masks perform comparably to ground truth in the context of classification under spurious correlations. Therefore, we consider the masks chosen by the \highlightblue{Class-Aided} foreground detector to be the closest available approximation to ground truth foreground masks in the absence of mask supervision.

\begin{table}
\vspace{-0.5cm}
    \centering
    \small
    \begin{tabular}{c|l}
\toprule
   FG Detector&WGA ($\uparrow$)\\
   \midrule
 -&83.6\\
 Max Prob&78.6\\ 
    Ens. $\mathcal{H}$&86.8\\
    Class-Aided&96.0\\ 
    Ground Truth&\textbf{96.7}\\
 \bottomrule
\end{tabular}
    \caption{\textbf{Different foreground detectors on Waterbirds} We report the worst-group accuracies on the Waterbirds dataset for different foreground detectors. Masks are generated by HQES and applied via ``Gray BG + Crop" (see §~\ref{sec:method:generate_representations}). The classification model is CLIP ViT-L-14 \citep{CLIP}. ``-" stands for classification of original images without using any masks. Max Prob stands for foreground detector that uses the following score (in terms of §~\ref{sec:method:robust_classifier}): $g_{\text{max\_prob}}(x, m) = \operatorname{max}_c p^c(\psi(a(x, m)))$ - maximum probability across all possible classes (its computation is equivalent to confidence in §~\ref{sec:ood_detection_main}). \highlightblue{Class-Aided} and \highlightpurple{Ens. $\mathcal{H}$} are described in §~\ref{sec:method:robust_classifier}. Ground Truth stands for ground truth foreground masks that are taken from \citep{DFR}.}
    \label{tab:non_fg}
\end{table}

\clearpage

\section{Extended implementation details}
\label{sec:impl}

\noindent\textbf{Classes for zero-shot classification} Following the original CLIP \citep{CLIP} work, we compute the classifier's pre-softmax logits $f(\psi(x))$ using dot products between image embeddings and text embeddings of class name prompts. Each prompt follows the format: ``A photo of $X$'', where $X$ is a class name from the corresponding dataset.

\noindent For Waterbirds \citep{Waterbirds} and UrbanCars \citep{Whaca}, we first compute dot products using prompts based on fine-grained class names from the Caltech Birds (CUB) dataset \citep{Wah2011TheCB} and the Stanford Cars dataset \citep{StanfordCars}, respectively. This is because the foreground objects in these datasets were originally cropped from the corresponding source datasets.

\noindent All fine-grained classes are then grouped into two broader categories. For Waterbirds, the classes are divided into ``land'' and ``sea'' birds. For UrbanCars, they are grouped into ``urban'' and ``country'' cars. The final prediction corresponds to the group containing the fine-grained class with the highest dot product.
\vspace{0.5em}

\noindent\textbf{How resize is done for \cropresize}
We apply the following steps to perform the \cropresize{} operation: (1) Find the smallest rectangle that fully contains the foreground object. (2) Expand the shorter side of this rectangle to match the longer side, ensuring that the center of the new square matches the original rectangle's center. (3) Resize the resulting square to the target resolution.

\vspace{0.5em}

\noindent\textbf{Fixed number of slots in OCL method}
When using FT-Dinosaur \citep{ft-dinosaur} as a mask generator in the OCL method, we fix the number of slots to $5$, following the recommendation from the original implementation.

\vspace{0.5em}

\noindent\textbf{Foundational segmentation model choice}
While HQES and SAM generally perform similarly on segmentation tasks, SAM shows significantly better performance on the mBO metric. Despite this, we use HQES in all of our main experiments, as we have full knowledge of its training data and can confirm that it was not trained on any of the datasets containing spurious correlations used in our evaluation.

\vspace{0.5em}

\noindent\textbf{Mask-free AlphaCLIP}
AlphaCLIP \citep{AlphaClip} requires a foreground mask as input. To simulate a mask-free setting, we use a mask that covers the entire image, effectively setting $\alpha = 1$. Although a mask is technically provided, it does not contain any useful localization information, so we treat this setup as mask-free performance.

\vspace{0.5em}

\noindent\textbf{Masks filtering} Before using masks in our experiments, we apply the following filtering rules:

\begin{enumerate}
    \item \textbf{Size:} Remove masks that cover less than $0.001$ of the image pixels.
    \item \textbf{Connected components:} Remove masks that contain more than $30$ connected components.
    \item \textbf{Background heuristic:} Remove masks that cover at least $6$ of the $8$ key points (the $4$ corners and the $4$ side centers of the image).
\end{enumerate}

\clearpage

\section{Additional details on FG detectors comparison}
\label{subsec:ood_detection}

In this section, we give additional details on comparing different candidates for FG detector methods apart from $g_{\text{class\_aided}}$ and 
$g_{\mathcal{H}}$ (see §~\ref{sec:method:robust_classifier} for details).

\vspace{0.5em}

\noindent\textbf{Dataset construction details.} We construct a binary classification dataset using the ImageNet validation set \citep{ImageNet}, considering only images that have ground truth bounding boxes for the main object (i.e., the one corresponding to the ground truth label). For each such image, we predict masks for all objects it contains, as described in the ``Generating masks'' paragraph in~§\ref{sec:method:generate_representations}. We then apply each mask using the \cropresize{} operation, following the ``Applying masks'' paragraph in~§\ref{sec:method:generate_representations}. Each resulting masked image is assigned a label as follows:

\vspace{0.2em}
\begin{itemize}[leftmargin=1.5em]
    \item Class $1$ (foreground) if its corresponding mask has the highest Intersection over Union (IoU; \citep{generalizedIoU}) with the ground truth bounding box.
    \item Class $0$ (non-foreground) otherwise.
\end{itemize}

\vspace{0.5em}

\noindent\textbf{How are OOD detectors used?} OOD detectors are used in the following way: First, we compute an uncertainty score for each sample using formulas from §~\ref{sec:ood_detection_main} based on the ensemble’s outputs (for single model entropy and \highlightpurple{Ens. $\mathcal{H}$} we additionally multiply this score by $-1$ so that it is lower for OOD samples than for ID samples). Then, we treat this uncertainty score as the probability of predicting class $1$ in our binary classification setting.

\noindent \underline{Note:} \highlightpurple{Ens. $\mathcal{H}$} corresponds to $g_{\mathcal{H}}$ and \highlightblue{Class-Aided} corresponds to $g_{\text{class\_aided}}$, as described in the ``Foreground detector'' paragraph in §~\ref{sec:method:robust_classifier}.

\vspace{0.5em}

\noindent\textbf{Ensemble members.} All model checkpoints are sourced from the ``openclip'' library \citep{openclip}, using the following pre-training dataset identifiers: ``openai'', ``datacomp\_xl\_s13b\_b90k'', ``dfn2b'', ``laion400m\_e31'', and ``laion400m\_e32''.

\noindent We focus on ensemble-based baselines for OOD detection, as they are among the most competitive approaches for this task \citep{Mukhoti2021DeepDU, ovadia2019can}.

%% file: main.bbl
\begin{thebibliography}{82}
\providecommand{\natexlab}[1]{#1}
\providecommand{\url}[1]{\texttt{#1}}
\expandafter\ifx\csname urlstyle\endcsname\relax
  \providecommand{\doi}[1]{doi: #1}\else
  \providecommand{\doi}{doi: \begingroup \urlstyle{rm}\Url}\fi

\bibitem[Arefin et~al.(2024)Arefin, Zhang, Baratin, Locatello, Rish, Liu, and Kawaguchi]{CoBalT}
Md~Rifat Arefin, Yan Zhang, Aristide Baratin, Francesco Locatello, Irina Rish, Dianbo Liu, and Kenji Kawaguchi.
\newblock Unsupervised concept discovery mitigates spurious correlations.
\newblock In \emph{International Conference on Machine Learning (ICML)}, 2024.

\bibitem[Asgari et~al.(2022)Asgari, Khani, Khani, Gholami, Tran, Mahdavi-Amiri, and Hamarneh]{MASKTUNE}
Saeid Asgari, Aliasghar Khani, Fereshte Khani, Ali Gholami, Linh Tran, Ali Mahdavi-Amiri, and Ghassan Hamarneh.
\newblock Masktune: Mitigating spurious correlations by forcing to explore.
\newblock In \emph{Advances in Neural Information Processing Systems}, 2022.

\bibitem[Ayzenberg and Behrmann(2024)]{ayzenberg2024development}
Vladislav Ayzenberg and Marlene Behrmann.
\newblock Development of visual object recognition.
\newblock \emph{Nature Reviews Psychology}, 3\penalty0 (2):\penalty0 73--90, 2024.

\bibitem[Berner et~al.(2019)Berner, Brockman, Chan, Cheung, Debiak, Dennison, Farhi, Fischer, Hashme, Hesse, J{\'o}zefowicz, Gray, Olsson, Pachocki, Petrov, de~Oliveira~Pinto, Raiman, Salimans, Schlatter, Schneider, Sidor, Sutskever, Tang, Wolski, and Zhang]{Berner2019Dota2W}
Christopher Berner, Greg Brockman, Brooke Chan, Vicki Cheung, Przemyslaw Debiak, Christy Dennison, David Farhi, Quirin Fischer, Shariq Hashme, Christopher Hesse, Rafal J{\'o}zefowicz, Scott Gray, Catherine Olsson, Jakub~W. Pachocki, Michael Petrov, Henrique~Pond{\'e} de Oliveira~Pinto, Jonathan Raiman, Tim Salimans, Jeremy Schlatter, Jonas Schneider, Szymon Sidor, Ilya Sutskever, Jie Tang, Filip Wolski, and Susan Zhang.
\newblock Dota 2 with large scale deep reinforcement learning.
\newblock \emph{ArXiv}, abs/1912.06680, 2019.

\bibitem[Brady et~al.(2023)Brady, Zimmermann, Sharma, Sch{\"o}lkopf, and von K{\"u}gelgen]{brady2023provably}
Jack Brady, Roland~S. Zimmermann, Yash Sharma, Bernhard Sch{\"o}lkopf, and Wieland~and von K{\"u}gelgen, Julius~Brendel.
\newblock Provably learning object-centric representations.
\newblock In \emph{International Conference on Machine Learning (ICML)}, 2023.

\bibitem[Brady et~al.(2024)Brady, von K{\"u}gelgen, Lachapelle, Buchholz, Kipf, and Brendel]{brady2024interaction}
Jack Brady, Julius von K{\"u}gelgen, S{\'e}bastien Lachapelle, Simon Buchholz, Thomas Kipf, and Wieland Brendel.
\newblock Interaction asymmetry: A general principle for learning composable abstractions.
\newblock \emph{arXiv preprint arXiv:2411.07784}, 2024.

\bibitem[Burgess et~al.(2019)Burgess, Matthey, Watters, Kabra, Higgins, Botvinick, and Lerchner]{MONET}
Christopher~P. Burgess, Lo{\"i}c Matthey, Nicholas Watters, Rishabh Kabra, Irina Higgins, Matthew~M. Botvinick, and Alexander Lerchner.
\newblock Monet: Unsupervised scene decomposition and representation.
\newblock \emph{ArXiv}, abs/1901.11390, 2019.

\bibitem[Caron et~al.(2021)Caron, Touvron, Misra, J'egou, Mairal, Bojanowski, and Joulin]{DINO}
Mathilde Caron, Hugo Touvron, Ishan Misra, Herv'e J'egou, Julien Mairal, Piotr Bojanowski, and Armand Joulin.
\newblock Emerging properties in self-supervised vision transformers.
\newblock \emph{International Conference on Computer Vision (ICCV)}, 2021.

\bibitem[Depeweg et~al.(2017)Depeweg, Hern{\'a}ndez-Lobato, Doshi-Velez, and Udluft]{Depeweg2017DecompositionOU}
Stefan Depeweg, Jos{\'e}~Miguel Hern{\'a}ndez-Lobato, Finale Doshi-Velez, and Steffen Udluft.
\newblock Decomposition of uncertainty in bayesian deep learning for efficient and risk-sensitive learning.
\newblock In \emph{International Conference on Machine Learning}, 2017.

\bibitem[Didolkar et~al.(2023)Didolkar, Goyal, and Bengio]{didolkar2023cycle}
Aniket Didolkar, Anirudh Goyal, and Yoshua Bengio.
\newblock Cycle consistency driven object discovery.
\newblock \emph{arXiv preprint arXiv:2306.02204}, 2023.

\bibitem[Didolkar et~al.(2025)Didolkar, Zadaianchuk, Goyal, Mozer, Bengio, Martius, and Seitzer]{ft-dinosaur}
Aniket Didolkar, Andrii Zadaianchuk, Anirudh Goyal, Mike Mozer, Yoshua Bengio, Georg Martius, and Maximilian Seitzer.
\newblock On the transfer of object-centric representation learning.
\newblock In \emph{International Conference on Learning Representations (ICLR)}, 2025.

\bibitem[Dittadi et~al.(2022)Dittadi, Papa, De~Vita, Sch{\"o}lkopf, Winther, and Locatello]{OCL-OOD-GEN-SEGMENT}
Andrea Dittadi, Samuele~S Papa, Michele De~Vita, Bernhard Sch{\"o}lkopf, Ole Winther, and Francesco Locatello.
\newblock Generalization and robustness implications in object-centric learning.
\newblock In \emph{International Conference on Machine Learning (ICML)}, 2022.

\bibitem[Dosovitskiy et~al.(2021)Dosovitskiy, Beyer, Kolesnikov, Weissenborn, Zhai, Unterthiner, Dehghani, Minderer, Heigold, Gelly, Uszkoreit, and Houlsby]{ViT}
Alexey Dosovitskiy, Lucas Beyer, Alexander Kolesnikov, Dirk Weissenborn, Xiaohua Zhai, Thomas Unterthiner, Mostafa Dehghani, Matthias Minderer, Georg Heigold, Sylvain Gelly, Jakob Uszkoreit, and Neil Houlsby.
\newblock An image is worth 16x16 words: Transformers for image recognition at scale.
\newblock In \emph{International Conference on Learning Representations (ICLR)}, 2021.

\bibitem[El-Nouby et~al.(2018)El-Nouby, Sharma, Schulz, Hjelm, Asri, Kahou, Bengio, and W.Taylor]{ElNouby2018TellDA}
Alaaeldin El-Nouby, Shikhar Sharma, Hannes Schulz, Devon Hjelm, Layla~El Asri, Samira~Ebrahimi Kahou, Yoshua Bengio, and Graham W.Taylor.
\newblock Tell, draw, and repeat: Generating and modifying images based on continual linguistic instruction.
\newblock \emph{2019 IEEE/CVF International Conference on Computer Vision (ICCV)}, pages 10303--10311, 2018.

\bibitem[Elsayed et~al.(2022)Elsayed, Mahendran, van Steenkiste, Greff, Mozer, and Kipf]{SAVI++}
Gamaleldin~F. Elsayed, Aravindh Mahendran, Sjoerd van Steenkiste, Klaus Greff, Michael~C. Mozer, and Thomas Kipf.
\newblock {SAVi++}: Towards end-to-end object-centric learning from real-world videos.
\newblock In \emph{Conference on Neural Information Processing Systems (NeurIPS)}, 2022.

\bibitem[Fumero et~al.(2023)Fumero, Wenzel, Zancato, Achille, Rodol\`{a}, Soatto, Sch\"{o}lkopf, and Locatello]{LatentFactorsRecoverRealWorld}
Marco Fumero, Florian Wenzel, Luca Zancato, Alessandro Achille, Emanuele Rodol\`{a}, Stefano Soatto, Bernhard Sch\"{o}lkopf, and Francesco Locatello.
\newblock Leveraging sparse and shared feature activations for disentangled representation learning.
\newblock In \emph{Advances in Neural Information Processing Systems}, pages 27682--27698. Curran Associates, Inc., 2023.

\bibitem[Greff et~al.(2019)Greff, Kaufman, Kabra, Watters, Burgess, Zoran, Matthey, Botvinick, and Lerchner]{greff2019multi}
Klaus Greff, Rapha{\"e}l~Lopez Kaufman, Rishabh Kabra, Nick Watters, Christopher Burgess, Daniel Zoran, Loic Matthey, Matthew Botvinick, and Alexander Lerchner.
\newblock Multi-object representation learning with iterative variational inference.
\newblock In \emph{International Conference on Machine Learning}, pages 2424--2433. PMLR, 2019.

\bibitem[Greff et~al.(2020)Greff, van Steenkiste, and Schmidhuber]{CompositionalNatureOfScenes}
Klaus Greff, Sjoerd van Steenkiste, and J{\"{u}}rgen Schmidhuber.
\newblock On the binding problem in artificial neural networks.
\newblock \emph{arXiv preprint arXiv:2012.05208}, 2020.

\bibitem[Greff et~al.(2022)Greff, Belletti, Beyer, Doersch, Du, Duckworth, Fleet, Gnanapragasam, Golemo, Herrmann, Kipf, Kundu, Lagun, Laradji, Liu, Meyer, Miao, Nowrouzezahrai, Oztireli, Pot, Radwan, Rebain, Sabour, Sajjadi, Sela, Sitzmann, Stone, Sun, Vora, Wang, Wu, Yi, Zhong, and Tagliasacchi]{Kubric}
Klaus Greff, Francois Belletti, Lucas Beyer, Carl Doersch, Yilun Du, Daniel Duckworth, David~J Fleet, Dan Gnanapragasam, Florian Golemo, Charles Herrmann, Thomas Kipf, Abhijit Kundu, Dmitry Lagun, Issam Laradji, Hsueh-Ti~(Derek) Liu, Henning Meyer, Yishu Miao, Derek Nowrouzezahrai, Cengiz Oztireli, Etienne Pot, Noha Radwan, Daniel Rebain, Sara Sabour, Mehdi S.~M. Sajjadi, Matan Sela, Vincent Sitzmann, Austin Stone, Deqing Sun, Suhani Vora, Ziyu Wang, Tianhao Wu, Kwang~Moo Yi, Fangcheng Zhong, and Andrea Tagliasacchi.
\newblock Kubric: a scalable dataset generator.
\newblock In \emph{Conference on Computer Vision and Pattern Recognition (CVPR)}, 2022.

\bibitem[Gruber and Buettner(2022)]{Gruber2022UncertaintyEO}
Sebastian Gruber and Florian Buettner.
\newblock Uncertainty estimates of predictions via a general bias-variance decomposition.
\newblock In \emph{International Conference on Artificial Intelligence and Statistics (AISTATS)}, 2022.

\bibitem[He et~al.(2016)He, Zhang, Ren, and Sun]{ResNet}
Kaiming He, Xiangyu Zhang, Shaoqing Ren, and Jian Sun.
\newblock Deep residual learning for image recognition.
\newblock In \emph{Conference on Computer Vision and Pattern Recognition (CVPR)}, 2016.

\bibitem[Hendrycks and Gimpel(2017)]{hendrycks2017a}
Dan Hendrycks and Kevin Gimpel.
\newblock A baseline for detecting misclassified and out-of-distribution examples in neural networks.
\newblock In \emph{International Conference on Learning Representations}, 2017.

\bibitem[Hubert and Arabie(1985)]{FG_ARI_1985}
Lawrence~J. Hubert and Phipps Arabie.
\newblock Comparing partitions.
\newblock \emph{Journal of Classification}, 1985.

\bibitem[Ilharco et~al.(2021)Ilharco, Wortsman, Wightman, Gordon, Carlini, Taori, Dave, Shankar, Namkoong, Miller, Hajishirzi, Farhadi, and Schmidt]{openclip}
Gabriel Ilharco, Mitchell Wortsman, Ross Wightman, Cade Gordon, Nicholas Carlini, Rohan Taori, Achal Dave, Vaishaal Shankar, Hongseok Namkoong, John Miller, Hannaneh Hajishirzi, Ali Farhadi, and Ludwig Schmidt.
\newblock Openclip.
\newblock In \emph{GitHub}. Zenodo, 2021.
\newblock If you use this software, please cite it as below.

\bibitem[Jiang et~al.(2023)Jiang, Deng, Singh, and Ahn]{SlotDiffusion}
Jindong Jiang, Fei Deng, Gautam Singh, and Sungjin Ahn.
\newblock Object-centric slot diffusion.
\newblock In \emph{Conference on Neural Information Processing Systems (NeurIPS)}, 2023.

\bibitem[Kapl et~al.(2025)Kapl, Mamaghan, Horn, Marr, Bauer, and Dittadi]{Ferdinand2}
Ferdinand Kapl, Amir Mohammad~Karimi Mamaghan, Max Horn, Carsten Marr, Stefan Bauer, and Andrea Dittadi.
\newblock Object-centric representations generalize better compositionally with less compute.
\newblock In \emph{Workshop on Spurious Correlation and Shortcut Learning: Foundations and Solutions}, 2025.

\bibitem[Karazija et~al.(2022)Karazija, Choudhury, Laina, Rupprecht, and Vedaldi]{karazija2022unsupervised}
Laurynas Karazija, Subhabrata Choudhury, Iro Laina, Christian Rupprecht, and Andrea Vedaldi.
\newblock Unsupervised multi-object segmentation by predicting probable motion patterns.
\newblock \emph{Advances in Neural Information Processing Systems}, 35:\penalty0 2128--2141, 2022.

\bibitem[Kipf et~al.(2022)Kipf, Elsayed, Mahendran, Stone, Sabour, Heigold, Jonschkowski, Dosovitskiy, and Greff]{SAVI}
Thomas Kipf, Gamaleldin~F. Elsayed, Aravindh Mahendran, Austin Stone, Sara Sabour, Georg Heigold, Rico Jonschkowski, Alexey Dosovitskiy, and Klaus Greff.
\newblock {Conditional Object-Centric Learning from Video}.
\newblock In \emph{International Conference on Learning Representations (ICLR)}, 2022.

\bibitem[Kirichenko et~al.(2023)Kirichenko, Izmailov, and Wilson]{DFR}
Polina Kirichenko, Pavel Izmailov, and Andrew~Gordon Wilson.
\newblock Last layer re-training is sufficient for robustness to spurious correlations.
\newblock In \emph{International Conference on Learning Representations (ICLR)}, 2023.

\bibitem[Kirillov et~al.(2023)Kirillov, Mintun, Ravi, Mao, Rolland, Gustafson, Xiao, Whitehead, Berg, Lo, Dollar, and Girshick]{SAM}
Alexander Kirillov, Eric Mintun, Nikhila Ravi, Hanzi Mao, Chloe Rolland, Laura Gustafson, Tete Xiao, Spencer Whitehead, Alexander~C. Berg, Wan-Yen Lo, Piotr Dollar, and Ross Girshick.
\newblock Segment anything.
\newblock In \emph{International Conference on Computer Vision (ICCV)}, 2023.

\bibitem[Kori et~al.(2023)Kori, Locatello, Ribeiro, Toni, and Glocker]{kori2023grounded}
Avinash Kori, Francesco Locatello, Fabio De~Sousa Ribeiro, Francesca Toni, and Ben Glocker.
\newblock Grounded object centric learning.
\newblock \emph{arXiv preprint arXiv:2307.09437}, 2023.

\bibitem[Krause et~al.(2013)Krause, Stark, Deng, and Fei-Fei]{StanfordCars}
Jonathan Krause, Michael Stark, Jia Deng, and Li Fei-Fei.
\newblock 3d object representations for fine-grained categorization.
\newblock \emph{2013 IEEE International Conference on Computer Vision Workshops}, pages 554--561, 2013.

\bibitem[Kulkarni et~al.(2019)Kulkarni, Gupta, Ionescu, Borgeaud, Reynolds, Zisserman, and Mnih]{Kulkarni2019UnsupervisedLO}
Tejas~D. Kulkarni, Ankush Gupta, Catalin Ionescu, Sebastian Borgeaud, Malcolm Reynolds, Andrew Zisserman, and Volodymyr Mnih.
\newblock Unsupervised learning of object keypoints for perception and control.
\newblock In \emph{Neural Information Processing Systems}, 2019.

\bibitem[Lakshminarayanan et~al.(2017)Lakshminarayanan, Pritzel, and Blundell]{Lakshminarayanan2017}
Balaji Lakshminarayanan, Alexander Pritzel, and Charles Blundell.
\newblock Simple and scalable predictive uncertainty estimation using deep ensembles.
\newblock In \emph{Advances in Neural Information Processing Systems}. Curran Associates, Inc., 2017.

\bibitem[Li et~al.(2023)Li, Evtimov, Gordo, Hazirbas, Hassner, Ferrer, Xu, and Ibrahim]{Whaca}
Zhiheng Li, Ivan Evtimov, Albert Gordo, Caner Hazirbas, Tal Hassner, Cristian~Canton Ferrer, Chenliang Xu, and Mark Ibrahim.
\newblock A whac-a-mole dilemma: Shortcuts come in multiples where mitigating one amplifies others.
\newblock In \emph{Conference on Computer Vision and Pattern Recognition (CVPR)}, 2023.

\bibitem[Liu et~al.(2021)Liu, Haghgoo, Chen, Raghunathan, Koh, Sagawa, Liang, and Finn]{JTT}
Evan~Z Liu, Behzad Haghgoo, Annie~S Chen, Aditi Raghunathan, Pang~Wei Koh, Shiori Sagawa, Percy Liang, and Chelsea Finn.
\newblock Just train twice: Improving group robustness without training group information.
\newblock In \emph{Proceedings of the 38th International Conference on Machine Learning}, pages 6781--6792. PMLR, 2021.

\bibitem[Liu et~al.(2023{\natexlab{a}})Liu, Li, Wu, and Lee]{LLAVA}
Haotian Liu, Chunyuan Li, Qingyang Wu, and Yong~Jae Lee.
\newblock Visual instruction tuning.
\newblock In \emph{Conference on Neural Information Processing Systems (NeurIPS)}, 2023{\natexlab{a}}.

\bibitem[Liu et~al.(2024{\natexlab{a}})Liu, Li, Li, and Lee]{LLAVA1_5}
Haotian Liu, Chunyuan Li, Yuheng Li, and Yong~Jae Lee.
\newblock Improved baselines with visual instruction tuning.
\newblock In \emph{Proceedings of the IEEE/CVF Conference on Computer Vision and Pattern Recognition (CVPR)}, pages 26296--26306, 2024{\natexlab{a}}.

\bibitem[Liu et~al.(2024{\natexlab{b}})Liu, Li, Li, Li, Zhang, Shen, and Lee]{LLAVA_NEXT}
Haotian Liu, Chunyuan Li, Yuheng Li, Bo Li, Yuanhan Zhang, Sheng Shen, and Yong~Jae Lee.
\newblock Llava-next: Improved reasoning, ocr, and world knowledge, 2024{\natexlab{b}}.

\bibitem[Liu et~al.(2023{\natexlab{b}})Liu, Alahi, Russell, Horn, Zietlow, Sch{\"o}lkopf, and Locatello]{liu2023causal}
Yuejiang Liu, Alexandre Alahi, Chris Russell, Max Horn, Dominik Zietlow, Bernhard Sch{\"o}lkopf, and Francesco Locatello.
\newblock Causal triplet: An open challenge for intervention-centric causal representation learning.
\newblock In \emph{Conference on Causal Learning and Reasoning}, pages 553--573. PMLR, 2023{\natexlab{b}}.

\bibitem[Locatello et~al.(2020)Locatello, Weissenborn, Unterthiner, Mahendran, Heigold, Uszkoreit, Dosovitskiy, and Kipf]{SlotAttention}
Francesco Locatello, Dirk Weissenborn, Thomas Unterthiner, Aravindh Mahendran, Georg Heigold, Jakob Uszkoreit, Alexey Dosovitskiy, and Thomas Kipf.
\newblock Object-centric learning with slot attention.
\newblock In \emph{Conference on Neural Information Processing Systems (NeurIPS)}, 2020.

\bibitem[Lu et~al.(2023)Lu, Kuen, Tiancheng, Jiuxiang, Weidong, Jiaya, Zhe, and Ming-Hsuan]{EntitySeg}
Qi Lu, Jason Kuen, Shen Tiancheng, Gu Jiuxiang, Guo Weidong, Jia Jiaya, Lin Zhe, and Yang Ming-Hsuan.
\newblock High-quality entity segmentation.
\newblock In \emph{International Conference on Computer Vision (ICCV)}, 2023.

\bibitem[Mamaghan et~al.(2025)Mamaghan, Papa, Johansson, Bauer, and Dittadi]{Ferdinand1}
Amir Mohammad~Karimi Mamaghan, Samuele Papa, Karl~Henrik Johansson, Stefan Bauer, and Andrea Dittadi.
\newblock Exploring the effectiveness of object-centric representations in visual question answering: Comparative insights with foundation models.
\newblock In \emph{International Conference on Learning Representations (ICLR)}, 2025.

\bibitem[Matsumori et~al.(2021)Matsumori, Shingyouchi, Abe, Fukuchi, Sugiura, and Imai]{Matsumori2021UnifiedQT}
Shoya Matsumori, Kosuke Shingyouchi, Yukikoko Abe, Yosuke Fukuchi, Komei Sugiura, and Michita Imai.
\newblock Unified questioner transformer for descriptive question generation in goal-oriented visual dialogue.
\newblock \emph{2021 IEEE/CVF International Conference on Computer Vision (ICCV)}, pages 1878--1887, 2021.

\bibitem[Migimatsu and Bohg(2020)]{migimatsu2019objectcentric}
Toki Migimatsu and Jeannette Bohg.
\newblock Object-centric task and motion planning in dynamic environments.
\newblock \emph{IEEE Robotics and Automation Letters}, 5\penalty0 (2):\penalty0 844--851, 2020.

\bibitem[Mucs{\'a}nyi et~al.(2024)Mucs{\'a}nyi, Kirchhof, and Oh]{BalintUncDecomp}
B{\'a}lint Mucs{\'a}nyi, Michael Kirchhof, and Seong~Joon Oh.
\newblock Benchmarking uncertainty disentanglement: Specialized uncertainties for specialized tasks.
\newblock In \emph{The Thirty-eight Conference on Neural Information Processing Systems Datasets and Benchmarks Track}, 2024.

\bibitem[Mukhoti et~al.(2021)Mukhoti, Kirsch, van Amersfoort, Torr, and Gal]{Mukhoti2021DeepDU}
Jishnu Mukhoti, Andreas Kirsch, Joost~R. van Amersfoort, Philip H.~S. Torr, and Yarin Gal.
\newblock Deep deterministic uncertainty: A new simple baseline.
\newblock \emph{Conference on Computer Vision and Pattern Recognition (CVPR)}, 2021.

\bibitem[Nam et~al.(2020)Nam, Cha, Ahn, Lee, and Shin]{LfF2020}
Junhyun Nam, Hyuntak Cha, Sungsoo Ahn, Jaeho Lee, and Jinwoo Shin.
\newblock Learning from failure: Training debiased classifier from biased classifier.
\newblock In \emph{Proceedings of the 34th International Conference on Neural Information Processing Systems}, Red Hook, NY, USA, 2020. Curran Associates Inc.

\bibitem[Ovadia et~al.(2019)Ovadia, Fertig, Ren, Nado, Sculley, Nowozin, Dillon, Lakshminarayanan, and Snoek]{ovadia2019can}
Yaniv Ovadia, Emily Fertig, Jie Ren, Zachary Nado, David Sculley, Sebastian Nowozin, Joshua Dillon, Balaji Lakshminarayanan, and Jasper Snoek.
\newblock Can you trust your model's uncertainty? evaluating predictive uncertainty under dataset shift.
\newblock \emph{Advances in neural information processing systems}, 32, 2019.

\bibitem[Pont-Tuset et~al.(2015)Pont-Tuset, Arbel{\'a}ez, Barron, Marqu{\'e}s, and Malik]{mbo_metric}
Jordi Pont-Tuset, Pablo Arbel{\'a}ez, Jonathan~T. Barron, Ferran Marqu{\'e}s, and Jitendra Malik.
\newblock Multiscale combinatorial grouping for image segmentation and object proposal generation.
\newblock \emph{Transactions on Pattern Analysis and Machine Intelligence (TPAMI)}, 2015.

\bibitem[Qiu et~al.(2023)Qiu, Potapczynski, Izmailov, and Wilson]{AFR}
Shikai Qiu, Andres Potapczynski, Pavel Izmailov, and Andrew~Gordon Wilson.
\newblock Simple and fast group robustness by automatic feature reweighting.
\newblock In \emph{Proceedings of the 40th International Conference on Machine Learning}. JMLR.org, 2023.

\bibitem[Radford et~al.(2021)Radford, Kim, Hallacy, Ramesh, Goh, Agarwal, Sastry, Askell, Mishkin, Clark, Krueger, and Sutskever]{CLIP}
Alec Radford, Jong~Wook Kim, Chris Hallacy, Aditya Ramesh, Gabriel Goh, Sandhini Agarwal, Girish Sastry, Amanda Askell, Pamela Mishkin, Jack Clark, Gretchen Krueger, and Ilya Sutskever.
\newblock Learning transferable visual models from natural language supervision.
\newblock In \emph{International Conference on Machine Learning (ICML)}, 2021.

\bibitem[Rand(1971)]{FG_ARI_1971}
William~M. Rand.
\newblock Objective criteria for the evaluation of clustering methods.
\newblock \emph{Journal of the American Statistical Association}, 1971.

\bibitem[Ravi et~al.(2025)Ravi, Gabeur, Hu, Hu, Ryali, Ma, Khedr, R{\"a}dle, Rolland, Gustafson, Mintun, Pan, Alwala, Carion, Wu, Girshick, Dollar, and Feichtenhofer]{SAMv2}
Nikhila Ravi, Valentin Gabeur, Yuan-Ting Hu, Ronghang Hu, Chaitanya Ryali, Tengyu Ma, Haitham Khedr, Roman R{\"a}dle, Chloe Rolland, Laura Gustafson, Eric Mintun, Junting Pan, Kalyan~Vasudev Alwala, Nicolas Carion, Chao-Yuan Wu, Ross Girshick, Piotr Dollar, and Christoph Feichtenhofer.
\newblock {SAM} 2: Segment anything in images and videos.
\newblock In \emph{International Conference on Learning Representations (ICLR)}, 2025.

\bibitem[Rezatofighi et~al.(2019)Rezatofighi, Tsoi, Gwak, Sadeghian, Reid, and Savarese]{generalizedIoU}
Seyed~Hamid Rezatofighi, Nathan Tsoi, JunYoung Gwak, Amir Sadeghian, Ian~D. Reid, and Silvio Savarese.
\newblock Generalized intersection over union: A metric and a loss for bounding box regression.
\newblock \emph{2019 IEEE/CVF Conference on Computer Vision and Pattern Recognition (CVPR)}, pages 658--666, 2019.

\bibitem[Rombach et~al.(2022)Rombach, Blattmann, Lorenz, Esser, and Ommer]{StableDiffusion}
Robin Rombach, Andreas Blattmann, Dominik Lorenz, Patrick Esser, and BjÃ¶rn Ommer.
\newblock High-resolution image synthesis with latent diffusion models.
\newblock In \emph{Conference on Computer Vision and Pattern Recognition (CVPR)}, 2022.

\bibitem[Rubinstein et~al.(2024)Rubinstein, Scimeca, Teney, and Oh]{rubinstein2024scalable}
Alexander Rubinstein, Luca Scimeca, Damien Teney, and Seong~Joon Oh.
\newblock Scalable ensemble diversification for ood generalization and detection.
\newblock \emph{arXiv preprint arXiv:2409.16797}, 2024.

\bibitem[Russakovsky et~al.(2015)Russakovsky, Deng, Su, Krause, Satheesh, Ma, Huang, Karpathy, Khosla, Bernstein, et~al.]{ImageNet}
Olga Russakovsky, Jia Deng, Hao Su, Jonathan Krause, Sanjeev Satheesh, Sean Ma, Zhiheng Huang, Andrej Karpathy, Aditya Khosla, Michael Bernstein, et~al.
\newblock Imagenet large scale visual recognition challenge.
\newblock \emph{International Journal of Computer Vision (IJCV)}, 2015.

\bibitem[Sagawa et~al.(2020)Sagawa, Koh, Hashimoto, and Liang]{Waterbirds}
Shiori Sagawa, Pang~Wei Koh, Tatsunori~B. Hashimoto, and Percy Liang.
\newblock Distributionally robust neural networks.
\newblock In \emph{International Conference on Learning Representations (ICLR)}, 2020.

\bibitem[Sauer and Geiger(2021)]{SIN}
Axel Sauer and Andreas Geiger.
\newblock Counterfactual generative networks.
\newblock In \emph{International Conference on Learning Representations}, 2021.

\bibitem[Sch{\"o}lkopf et~al.(2021)Sch{\"o}lkopf, Locatello, Bauer, Ke, Kalchbrenner, Goyal, and Bengio]{ScholkopfCausal}
B. Sch{\"o}lkopf, F. Locatello, S. Bauer, N.~R. Ke, N. Kalchbrenner, A. Goyal, and Y. Bengio.
\newblock Toward causal representation learning.
\newblock \emph{Proceedings of the IEEE}, 2021.

\bibitem[Seitzer et~al.(2023)Seitzer, Horn, Zadaianchuk, Zietlow, Xiao, Simon-Gabriel, He, Zhang, Sch{\"o}lkopf, Brox, and Locatello]{Dinosaur}
Maximilian Seitzer, Max Horn, Andrii Zadaianchuk, Dominik Zietlow, Tianjun Xiao, Carl-Johann Simon-Gabriel, Tong He, Zheng Zhang, Bernhard Sch{\"o}lkopf, Thomas Brox, and Francesco Locatello.
\newblock Bridging the gap to real-world object-centric learning.
\newblock In \emph{International Conference on Learning Representations (ICLR)}, 2023.

\bibitem[Spelke(1990{\natexlab{a}})]{CognitionSpelke}
Elizabeth~S. Spelke.
\newblock Principles of object perception.
\newblock \emph{Cognitive Science}, 1990{\natexlab{a}}.

\bibitem[Spelke(1990{\natexlab{b}})]{spelke1990principles}
Elizabeth~S Spelke.
\newblock Principles of object perception.
\newblock \emph{Cognitive science}, 1990{\natexlab{b}}.

\bibitem[Sun et~al.(2019)Sun, Karlsson, Wu, Tenenbaum, and Murphy]{sun2018predicting}
Chen Sun, Per Karlsson, Jiajun Wu, Joshua~B Tenenbaum, and Kevin Murphy.
\newblock Predicting the present and future states of multi-agent systems from partially-observed visual data.
\newblock In \emph{International Conference on Learning Representations}, 2019.

\bibitem[Sun et~al.(2023)Sun, Fang, Wu, Zhang, Zang, Kong, Xiong, Lin, and Wang]{AlphaClip}
Zeyi Sun, Ye Fang, Tong Wu, Pan Zhang, Yuhang Zang, Shu Kong, Yuanjun Xiong, Dahua Lin, and Jiaqi Wang.
\newblock Alpha-clip: A clip model focusing on wherever you want.
\newblock \emph{Conference on Computer Vision and Pattern Recognition (CVPR)}, 2023.

\bibitem[Taghanaki et~al.(2021)Taghanaki, Choi, Khasahmadi, and Goyal]{CIM}
Saeid~Asgari Taghanaki, Kristy Choi, Amir~Hosein Khasahmadi, and Anirudh Goyal.
\newblock Robust representation learning via perceptual similarity metrics.
\newblock In \emph{International Conference on Machine Learning (ICML)}, 2021.

\bibitem[Tran et~al.(2022)Tran, Liu, Dusenberry, Phan, Collier, Ren, Han, Wang, Mariet, Hu, Band, Rudner, Singhal, Nado, van Amersfoort, Kirsch, Jenatton, Thain, Yuan, Buchanan, Murphy, Sculley, Gal, Ghahramani, Snoek, and Lakshminarayanan]{Tran2022PlexTR}
Dustin Tran, Jeremiah~Zhe Liu, Michael~W. Dusenberry, Du Phan, Mark Collier, Jie~Jessie Ren, Kehang Han, Z. Wang, Zelda~E. Mariet, Huiyi Hu, Neil Band, Tim G.~J. Rudner, K. Singhal, Zachary Nado, Joost~R. van Amersfoort, Andreas Kirsch, Rodolphe Jenatton, Nithum Thain, Honglin Yuan, E.~Kelly Buchanan, Kevin Murphy, D. Sculley, Yarin Gal, Zoubin Ghahramani, Jasper Snoek, and Balaji Lakshminarayanan.
\newblock Plex: Towards reliability using pretrained large model extensions.
\newblock \emph{First Workshop on Pre-training: Perspectives, Pitfalls, and Paths Forward at ICML 2022}, 2022.

\bibitem[Téglás et~al.(2011)Téglás, Vul, Girotto, Gonzalez, Tenenbaum, and Bonatti]{InfantsCognition}
Ernő Téglás, Edward Vul, Vittorio Girotto, Michel Gonzalez, Joshua~B. Tenenbaum, and Luca~L. Bonatti.
\newblock Pure reasoning in 12-month-old infants as probabilistic inference.
\newblock \emph{Science}, 2011.

\bibitem[Wagemans(2015)]{CognitionOxford}
Johan Wagemans.
\newblock \emph{The Oxford Handbook of Perceptual Organization}.
\newblock Oxford University Press, 2015.

\bibitem[Wah et~al.(2011)Wah, Branson, Welinder, Perona, and Belongie]{Wah2011TheCB}
Catherine Wah, Steve Branson, Peter Welinder, Pietro Perona, and Serge~J. Belongie.
\newblock The caltech-ucsd birds-200-2011 dataset.
\newblock In \emph{Technical report in California Institute of Technology}, 2011.

\bibitem[Wang et~al.(2024)Wang, Lin, Chen, Schmidt, Han, and Zhang]{CounterAnimals}
Qizhou Wang, Yong Lin, Yongqiang Chen, Ludwig Schmidt, Bo Han, and Tong Zhang.
\newblock A sober look at the robustness of clips to spurious features.
\newblock In \emph{Conference on Neural Information Processing Systems (NeurIPS)}, 2024.

\bibitem[Watters et~al.(2019)Watters, Matthey, Bosnjak, Burgess, and Lerchner]{Watters2019COBRADM}
Nicholas Watters, Lo{\"i}c Matthey, Matko Bosnjak, Christopher~P. Burgess, and Alexander Lerchner.
\newblock Cobra: Data-efficient model-based rl through unsupervised object discovery and curiosity-driven exploration.
\newblock \emph{ArXiv}, abs/1905.09275, 2019.

\bibitem[Webb et~al.(2023)Webb, Mondal, and Cohen]{webb2023systematic}
Taylor~Whittington Webb, Shanka~Subhra Mondal, and Jonathan Cohen.
\newblock Systematic visual reasoning through object-centric relational abstraction.
\newblock In \emph{Thirty-seventh Conference on Neural Information Processing Systems}, 2023.

\bibitem[Wiedemer et~al.(2024)Wiedemer, Brady, Panfilov, Juhos, Bethge, and Brendel]{wiedemer2024provable}
Thadd{\"a}us Wiedemer, Jack Brady, Alexander Panfilov, Attila Juhos, Matthias Bethge, and Wieland Brendel.
\newblock Provable compositional generalization for object-centric learning.
\newblock In \emph{International Conference on Learning Representations (ICLR)}, 2024.

\bibitem[Xiao et~al.(2021)Xiao, Engstrom, Ilyas, and Madry]{IN-9}
Kai~Yuanqing Xiao, Logan Engstrom, Andrew Ilyas, and Aleksander Madry.
\newblock Noise or signal: The role of image backgrounds in object recognition.
\newblock In \emph{International Conference on Learning Representations(ICLR)}, 2021.

\bibitem[Yang et~al.(2020)Yang, Mao, Wu, Parikh, Cox, Tenenbaum, and Gan]{Yang2020ObjectCentricDO}
Jianwei Yang, Jiayuan Mao, Jiajun Wu, Devi Parikh, David Cox, Joshua~B. Tenenbaum, and Chuang Gan.
\newblock Object-centric diagnosis of visual reasoning.
\newblock \emph{ArXiv}, abs/2012.11587, 2020.

\bibitem[Yang et~al.(2023)Yang, Gan, Dziugaite, and Mirzasoleiman]{SPARE}
Yu Yang, Eric Gan, Gintare~Karolina Dziugaite, and Baharan Mirzasoleiman.
\newblock Identifying spurious biases early in training through the lens of simplicity bias.
\newblock \emph{ArXiv}, abs/2305.18761, 2023.

\bibitem[Yoon et~al.(2023)Yoon, Wu, Bae, and Ahn]{yoon2023investigation}
Jaesik Yoon, Yi-Fu Wu, Heechul Bae, and Sungjin Ahn.
\newblock An investigation into pre-training object-centric representations for reinforcement learning.
\newblock In \emph{International Conference on Machine Learning}, pages 40147--40174. PMLR, 2023.

\bibitem[Zhai et~al.(2023)Zhai, Mustafa, Kolesnikov, and Beyer]{SigLip}
Xiaohua Zhai, Basil Mustafa, Alexander Kolesnikov, and Lucas Beyer.
\newblock Sigmoid loss for language image pre-training.
\newblock \emph{2023 IEEE/CVF International Conference on Computer Vision (ICCV)}, pages 11941--11952, 2023.

\bibitem[Zhang et~al.(2024)Zhang, Pan, Kim, Kweon, and Mao]{IN-D}
Chenshuang Zhang, Fei Pan, Junmo Kim, In~So Kweon, and Chengzhi Mao.
\newblock Imagenet-d: Benchmarking neural network robustness on diffusion synthetic object.
\newblock \emph{Conference on Computer Vision and Pattern Recognition (CVPR)}, 2024.

\bibitem[Zhu et~al.(2024)Zhu, Chen, Shen, Li, and Elhoseiny]{minigpt4}
Deyao Zhu, Jun Chen, Xiaoqian Shen, Xiang Li, and Mohamed Elhoseiny.
\newblock Mini{GPT}-4: Enhancing vision-language understanding with advanced large language models.
\newblock In \emph{The Twelfth International Conference on Learning Representations}, 2024.

\end{thebibliography}
